\documentclass[lettersize,journal]{IEEEtran}
\usepackage{amsmath,amsfonts}
\usepackage{algorithmic}
\usepackage{algorithm}
\usepackage{array}
\usepackage[caption=false,font=normalsize,labelfont=sf,textfont=sf]{subfig}
\usepackage{textcomp}
\usepackage{stfloats}
\usepackage{url}
\usepackage{verbatim}
\usepackage{graphicx}
\usepackage{cite}
\usepackage{booktabs}
\usepackage{algorithm}
\usepackage{algorithmic}
\usepackage[switch]{lineno}
\usepackage{bm}        
\usepackage{array}

\usepackage{placeins}
\usepackage{mathtools}

\usepackage{multirow}
\usepackage{makecell}
\usepackage{graphicx}
\usepackage[table]{xcolor}
\usepackage{lipsum} 
\usepackage{booktabs}
\usepackage{tikz}
\usepackage{tabularx}
\usepackage{newfloat}
\begin{document}

\title{Beyond Post-Quantization: Native Hash Learning\\with a Dedicated HASH Token}


\author{Xinze Liu, Ding Wang, Dayan Wu, Hengjie Zhu, Peng Fu, Zheng Lin and Weiping Wang%
\thanks{Corresponding author: Dayan Wu.}%
\thanks{Xinze Liu, Dayan Wu, Hengjie Zhu, Peng Fu and Zheng Lin are with the Institute of Information Engineering, CAS, Beijing 100190, China. Xinze Liu and Hengjie Zhu are also with the School of Cyber Security, University of Chinese Academy of Sciences, Beijing 100190, China (e-mail: liuxinze@iie.ac.cn; wudayan@iie.ac.cn; zhuhengjie@iie.ac.cn; fupeng@iie.ac.cn; linzheng@iie.ac.cn; wangweiping@iie.ac.cn).}%
\thanks{Ding Wang is with Department of Applied Mathematics and Statistics, Johns Hopkins University, Baltimore, Maryland 21218, USA (e-mail: dwang141@jh.edu).}%
}
\markboth{Preprint}{Liu \MakeLowercase{\textit{et al.}}: Native Hash Learning with a Dedicated HASH Token}



\maketitle

\begin{abstract}
Efficient large-scale image retrieval requires compact representations that preserve semantic similarity under fast Hamming-space search. Deep hashing provides an appealing solution, but most existing CNN- and ViT-based methods still follow a post-quantization paradigm, where continuous visual features are first learned and binary codes are then produced by a terminal hash projection or binarization operation. This late code generation creates a feature-to-code discrepancy between the continuously optimized representation space and the discrete Hamming space used for retrieval. To address this limitation, we propose HashViT, a Vision Transformer framework for native hash token learning. Instead of treating hashing as a terminal readout, HashViT introduces a dedicated HASH token that serves as a persistent, hash-oriented retrieval state inside the transformer.
The HASH token is structurally decomposed into a Hash Register for direct binary code generation and a Semantic Workspace for preserving auxiliary continuous semantics. To enable effective workspace-to-register interaction, we further design a lightweight Hash Refinement Adapter that progressively refines the Hash Register across transformer layers. As a result, binary-oriented representations are formed through token evolution within the backbone, rather than being abruptly induced by an output-level projection. HashViT is optimized with a unified objective that combines learnable semantic center supervision, class-token similarity distillation, and quantization regularization, encouraging the HASH token to encode semantically structured and compact binary representations. 
Extensive experiments on three widely used benchmarks demonstrate that HashViT achieves state-of-the-art or highly competitive retrieval performance while preserving the efficiency of compact Hamming codes.
The source code is available at https://github.com/Xinze919/HashViT.
\end{abstract}
\section{Introduction}
\IEEEPARstart{T}{he} rapid growth of multimedia data has made efficient and accurate image retrieval a fundamental requirement in large-scale visual applications~\cite{perronnin2010large,noh2017large,radenovic2018revisiting}. 
Among existing solutions, deep hashing methods~\cite{CNNH,DPSH,ADSH,DAPH,CSQ,wu2019deep,wu2023deep,su2024data,wu2024pairwise,DCAH} have attracted sustained attention because they represent images with compact binary codes, enabling low storage cost and efficient Hamming-distance-based search. 
Unlike traditional hashing methods based on hand-crafted features or data-independent projections~\cite{SH,LSH,ITQ}, deep hashing jointly learns visual representations and hash functions in an end-to-end manner, and has become a widely adopted paradigm for large-scale image retrieval. In particular, most existing methods focus on designing stronger similarity-preserving losses, quantization regularizers, or visual encoders, while leaving the architectural role of the hash code largely unchanged: the binary code is generated only after continuous visual representations have already been formed. This raises a fundamental question for deep hashing: should the hash code remain a terminal output of a generic feature extractor, or can it be modeled as a native retrieval state that evolves inside the network? Here, by native we mean that the binary-oriented state is carried and refined inside the backbone; it does not imply that the final sign operation is removed.

\begin{figure}
  \centering
  \includegraphics[width=\linewidth]{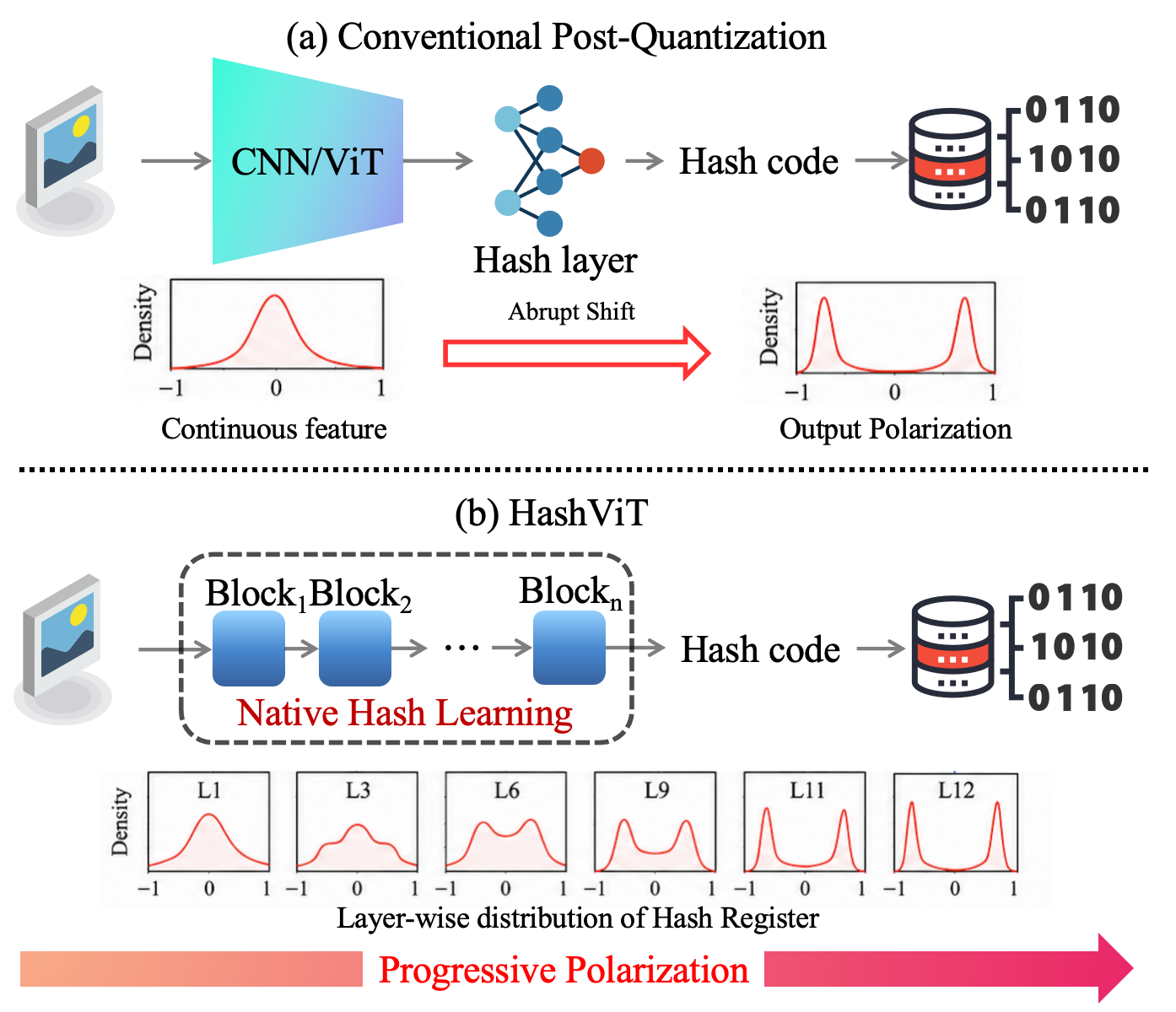}
    \caption{
    Motivation of HashViT.
    (a) Conventional post-quantization hashing first learns continuous backbone features and then maps them to polarized hash features through a terminal hash layer, causing an abrupt output-level transition from the continuous feature space to the Hamming space.
    (b) HashViT maintains a dedicated HASH token inside the transformer, whose Hash Register gradually evolves toward a polarized distribution across layers.
    }
  \label{motivation}
\end{figure}

Early deep hashing methods, such as HashNet~\cite{HashNet} and Deep Cauchy Hashing (DCH)~\cite{DCH}, demonstrated that learning hashing-oriented deep representations can significantly improve retrieval performance. 
Despite these advances, as illustrated in Fig.~\ref{motivation}(a), a dominant design in existing CNN- and ViT-based hashing methods remains post-quantization: the backbone first extracts continuous visual representations, and binary hash codes are generated only at the network output through a hash projection layer, a sign function, or an explicit quantization operation~\cite{HashNet,CSQ,HybridHash,MambaHash}. 
Under this paradigm, the hash code is not maintained as an internal representation during feature extraction, but is instead obtained as a terminal readout from an already formed continuous representation. Consequently, hashing is treated as an output-level discretization problem rather than as a representation-learning process that is intrinsically coupled with feature evolution.

Such late code generation leads to what we refer to as a feature-to-code discrepancy: a mismatch between the continuously optimized feature space and the discrete Hamming space, which we later operationalize through the layer-wise quantization error in Eq.~\ref{eq:layer}.
Intermediate representations are optimized mainly in a continuous feature space, whereas retrieval is eventually performed in the discrete Hamming space. 
As a result, the semantic structure learned by the backbone must be compressed into binary codes only at the final stage. 
Although stronger quantization constraints can partially alleviate this mismatch, enforcing aggressive binarization  directly on general-purpose visual features may impair their semantic expressiveness. 
This creates an inherent dilemma for deep hashing: late quantization weakens the coupling between representation learning and code formation, while indiscriminate intermediate quantization may interfere with continuous semantic modeling. This dilemma becomes more pronounced in transformer-based hashing. Vision Transformers are built upon progressive token interactions, where semantic information is gradually aggregated and refined across layers. However, existing hashing methods usually exploit this progressive evolution only for continuous representation learning, while binary code formation is still postponed to the final output stage. As a result, the Hamming-space structure required for retrieval is weakly involved in the internal representation dynamics of the backbone.

We argue that a key structural reason for this dilemma is the absence of a dedicated internal carrier for binary-oriented retrieval states. 
In conventional architectures, the same feature stream is expected to preserve rich continuous semantics and satisfy the discrete constraints required by hash codes. 
This design makes binary code learning largely dependent on the final projection head, and leaves the formation of hash representations weakly coupled with backbone feature learning. 
A more principled solution is to decouple the retrieval-specific binary state from the general semantic representation stream, so that continuous semantic modeling and hash code formation can be coordinated inside the network.

To this end, we propose HashViT, a native hash token learning framework built upon Vision Transformers. 
As shown in Fig.~\ref{motivation}(b), HashViT introduces a dedicated HASH token into the transformer sequence to serve as a persistent, hash-oriented retrieval state. 
Unlike conventional post-quantization methods that generate hash codes only at the output, the HASH token participates in self-attention throughout the network and evolves jointly with visual tokens. 
In this way, binary-oriented representations are formed through token evolution inside the transformer, rather than being induced only by a terminal hash projection. 
Therefore, the hash representation is no longer a terminal readout of the backbone: the HASH token attends to image tokens from the first layer and is refined after every block, and we empirically observe that its register progressively develops a binary-friendly, bimodal distribution across layers (Fig.~\ref{fig:layerwise_quantization}).

Although the HASH token is designed as the internal carrier of hash-related semantics, directly using only its first $B$ dimensions for binary code generation would discard the remaining dimensions that may still contain useful continuous semantic information. Meanwhile, enforcing binary constraints on the entire HASH token would be overly restrictive, as it may weaken the token's ability to preserve auxiliary semantics during transformer-layer evolution. To address this issue, we structurally decompose the HASH token into two complementary subspaces: a Hash Register and a Semantic Workspace. The Hash Register is aligned with the target Hamming space and is directly used for final binary code generation, while the Semantic Workspace preserves auxiliary continuous semantics that are not explicitly binarized. Based on this decomposition, we further introduce a lightweight Hash Refinement Adapter (HRA), which performs workspace-to-register refinement after each transformer block. In this way, the semantic information retained in the workspace can be progressively transferred to the Hash Register, allowing the binary-oriented retrieval state to remain explicit and persistent while still benefiting from continuous semantic modeling.

To further enhance semantic discriminability, we formulate a unified optimization objective for native HASH-token learning. 
The objective integrates learnable semantic center supervision, class-token similarity distillation, and quantization regularization. 
The learnable semantic centers provide class-level anchors in the hash space, the similarity distillation term preserves instance-level relational topology from the continuous class-token representation, and the quantization regularizer stabilizes the final binarization. 
By jointly optimizing these objectives, HashViT learns semantically structured, compact, and retrieval-effective binary representations inside the visual backbone. 
Together, the proposed architecture and objective offer a new formulation for deep hashing, in which binary code formation is explicitly represented, progressively refined, and semantically supervised within the ViT backbone, rather than appended as a final projection step.

The main contributions of this work are summarized as follows:

\begin{itemize}
\item We propose HashViT, a native hash representation learning framework for deep image hashing. Instead of treating hashing as a terminal readout, HashViT introduces a dedicated HASH token that maintains a hash-oriented retrieval state inside the Vision Transformer and evolves jointly with visual tokens throughout the backbone.
\item We structurally decompose the HASH token into a Hash Register and a Semantic Workspace. The Hash Register generates the binary code, while the Semantic Workspace preserves auxiliary continuous semantics and refines the register through a lightweight Hash Refinement Adapter (HRA), allowing hash representations to evolve within the backbone without directly quantizing the entire visual feature stream.

\item Finally, we train HashViT with a unified semantic supervision objective that combines learnable text-initialized semantic centers, class-token similarity distillation, and quantization regularization. Experiments on three benchmarks validate the effectiveness of the proposed framework.
\end{itemize}

The remainder of this paper is organized as follows. Section~\ref{sec:related_work} reviews related work. Section~\ref{sec:method} presents the proposed HashViT framework. Section~\ref{exp} reports experimental results and analyses. Section~\ref{conclusion} concludes the paper.

\section{Related Work}
\label{sec:related_work}

\subsection{Deep Image Hashing under the Post-Quantization}

Large-scale image retrieval aims to return semantically relevant images from a large database given a query image. Early retrieval methods mainly rely on hand-crafted visual descriptors or data-independent hashing functions~\cite{SH,LSH,ITQ}. Although these methods are computationally efficient, their representation capacity is limited, making it difficult to capture high-level semantic similarity in complex visual scenarios. Deep hashing has therefore become a widely adopted solution for efficient image retrieval~\cite{gu2022deep,wu2019deep,yang2020deep,su2024data,wu2024pairwise,wu2023deep,pushe,su2025boundary}, as it jointly learns discriminative image representations and compact binary codes in an end-to-end manner.

A large body of supervised deep hashing methods has been developed based on convolutional neural networks. DPSH~\cite{DPSH} jointly learns image features and hash codes from pairwise semantic similarities in an end-to-end framework. HashNet~\cite{HashNet} introduces a continuation strategy and a weighted pairwise cross-entropy loss to alleviate the optimization difficulty caused by the non-differentiable sign function and imbalanced similarity pairs. DCH~\cite{DCH} models pairwise similarity with a Cauchy-distribution-based objective, which imposes stronger penalties on similar image pairs with large Hamming distances. DPN~\cite{DPN} further proposes a polarization loss to encourage continuous outputs to approach binary values. For multi-label image retrieval, recent methods further exploit label correlations and neighborhood structures to improve the discriminability of binary embeddings. For example, Deep Neighbor Discriminant Binary Embedding~\cite{NDBE} enhances neighbor-level discrimination for multi-label image retrieval, showing the importance of preserving fine-grained semantic relations in the learned Hamming space. These methods effectively improve the discriminability of learned hash codes, but their binary representations are still mainly optimized at the output stage.

With the development of Vision Transformers, Transformer-based architectures have also been introduced into deep hashing. TransHash~\cite{transhash} presents a Transformer-based hashing framework and explores multi-granularity feature learning for image retrieval. MSViT~\cite{msvit} extracts multi-scale patch representations to enhance retrieval robustness, while HybridHash~\cite{HybridHash} combines CNNs and Transformers to jointly capture local structures and global dependencies. More recently, MambaHash~\cite{MambaHash} introduces visual state space models into deep hashing to improve long-range modeling and representation diversity. These methods demonstrate the advantage of modern visual backbones for hashing retrieval, but their binary codes are still usually produced by a final hash layer or an output-level quantization operation.

Despite their architectural differences, most existing CNN- and Transformer-based hashing methods share a common post-quantization paradigm. Specifically, the network first learns continuous visual features, and binary hash codes are generated only at the final stage through a sign function, a hash projection layer, or an output-level quantization constraint. This paradigm creates a discrepancy between the continuous feature space optimized during representation learning and the discrete Hamming space used during retrieval. Strengthening the final quantization constraint can reduce this discrepancy to some extent, but overly aggressive binarization may also damage semantic discrimination. Therefore, existing methods face an inherent tension between continuous semantic representation learning and discrete binary code formation.

Different from these methods that mainly optimize the final feature-to-code mapping, our work investigates how binary-oriented representations can be formed as an internal state of the visual backbone. HashViT addresses this problem by introducing a dedicated HASH token into the Vision Transformer, allowing hash representations to be progressively refined across transformer layers before the final sign operation.

\subsection{Hash Center Learning and Semantic Hash Centers}

Hash center learning has become an important direction in deep hashing because it provides explicit semantic structures in the Hamming space. Instead of only preserving pairwise similarities among samples, center-based methods assign each semantic category one or multiple representative hash centers and encourage image hash codes to be close to their corresponding centers. This formulation improves intra-class compactness and inter-class separability, and often leads to more stable optimization than purely pairwise objectives. However, most center-based methods still use centers as supervision targets for the final hash features, while the internal feature extraction process remains continuous.

Early center-based methods mainly focus on constructing well-separated hash centers in the Hamming space. CSQ~\cite{CSQ} introduces central similarity quantization and constructs hash centers with sufficiently large pairwise Hamming distances, typically using Hadamard matrices or random Bernoulli sampling. By pulling samples toward their corresponding centers, CSQ transforms supervised hashing into a classification-like learning problem in the Hamming space. OrthoHash~\cite{OrthoHash} further encourages continuous representations to align with class-specific orthogonal hash centers under cosine similarity. CenterHash~\cite{CenterHash} constructs minimal-distance-separated hash centers by explicitly optimizing pairwise distances among centers, providing stronger separation between different categories. These methods show that well-designed centers can significantly improve the structure of the learned Hamming space, but the centers are usually fixed before training and cannot adapt to the evolving visual representation distribution.
\begin{figure*}
  \centering
  \includegraphics[width=\linewidth]{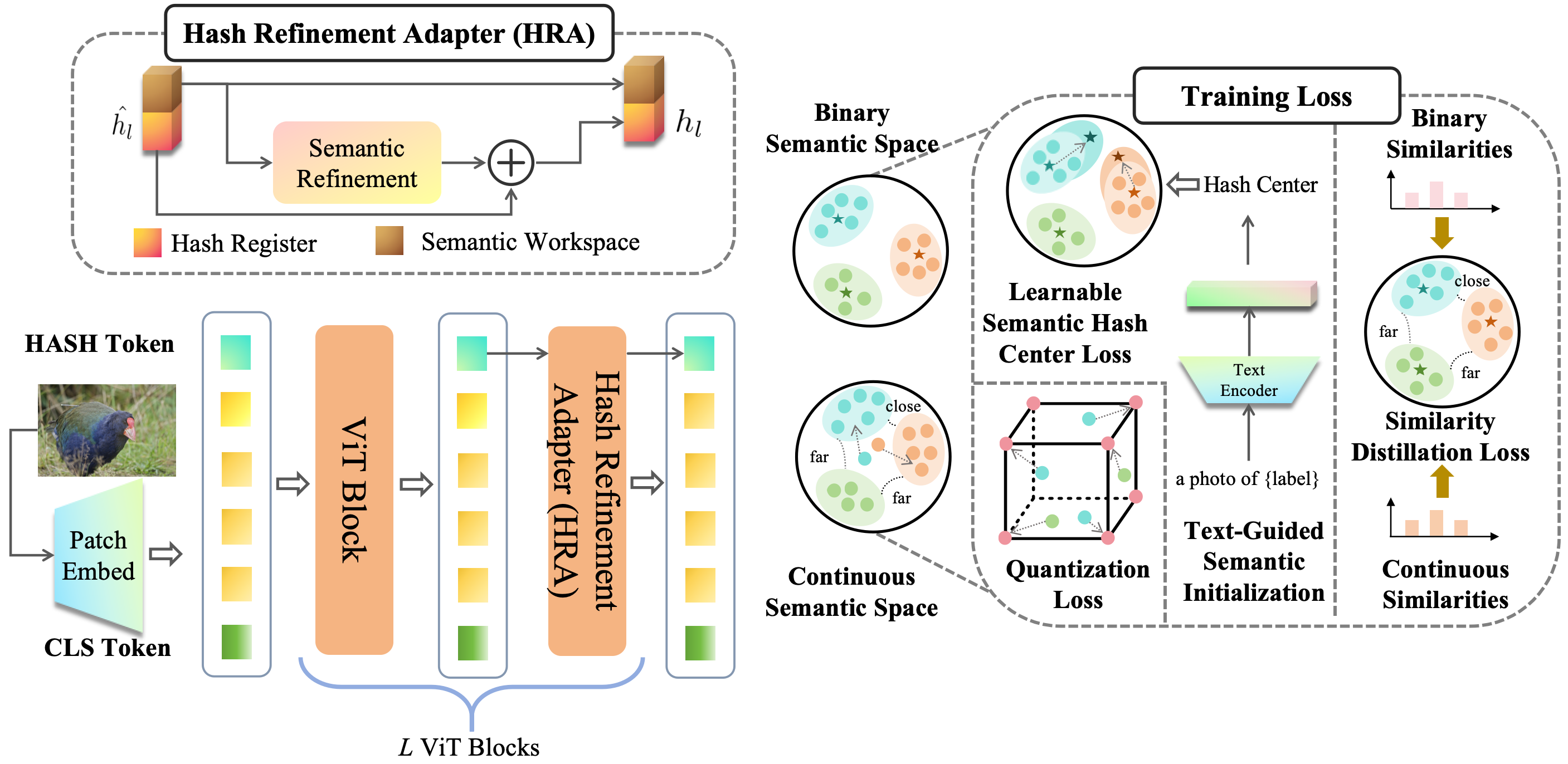}
    \caption{
    Overview of HashViT. The CLS token captures continuous visual semantics, while the HASH token learns native hash representations. Text-guided semantic hash centers, similarity distillation, and quantization regularization jointly supervise discriminative and binary-friendly hash learning.
    }
  \label{framework}
\end{figure*}
To overcome the rigidity of fixed centers, several works have explored learnable or dynamically updated hash centers. Learnable Central Similarity Quantization~\cite{LCSQ} extends the central similarity idea by introducing data-dependent learnable centers for efficient image and video retrieval. Deep Hashing with Hash Center Update~\cite{HCU} further studies dynamic hash center updating, allowing the target centers to better match the learned feature distribution. For multi-label retrieval, MCR~\cite{MCR} introduces a multi-central ranking loss with learnable hash centers and data-to-data similarities, alleviating the conflict caused by assigning complex multi-label samples to overly rigid proxy targets. These methods indicate that learnable centers are more flexible than fixed handcrafted centers. However, they still primarily improve the target structure of the final hash space, while binary codes are generated through output-level quantization.

Another important trend is to incorporate semantic relationships into hash center construction. Traditional center-based methods often assume that all different categories should be equally separated in the Hamming space. This assumption is too restrictive for multi-label or fine-grained retrieval, where different classes may have meaningful semantic correlations. Image Retrieval with Well-Separated Semantic Hash Centers~\cite{WSHC} attempts to construct centers by considering both semantic relationships and inter-center separation. SHC~\cite{SHC} further emphasizes that semantically related classes should have closer centers, while unrelated classes should remain far apart. More recently, Codebook-Centric Deep Hashing~\cite{CRH} studies the end-to-end joint learning of semantic hash centers and the neural hash function from a codebook-centric perspective, further showing that hash center assignment and representation learning should be optimized in a coordinated manner. These works improve the semantic organization of hash centers, but most of them still supervise the final continuous hash representation rather than changing the internal representation learning process.

Although learnable semantic hash centers improve the supervision of binary code learning, they do not fully resolve the feature-to-code mismatch caused by post-quantization. Even when the centers are adaptive or semantically structured, the network usually learns continuous features first and then projects them into binary codes at the end. As a result, the semantic center structure and the binary representation formation are still weakly coupled. HashViT complements this line of research by coupling semantic center supervision with a dedicated HASH token that evolves inside the Transformer. In this way, semantic centers not only guide the target structure of the Hamming space, but also supervise the progressive formation of binary-oriented representations inside the visual backbone.

\section{Proposed Method}   
\label{sec:method}
\subsection{Problem Formulation}
\label{subsec:problem}

Let $\mathcal{D}=\{(\mathbf{x}_i,\mathbf{y}_i)\}_{i=1}^{N}$ denote a training set, where $\mathbf{x}_i\in\mathbb{R}^{H\times W\times 3}$ is the $i$-th image and $\mathbf{y}_i\in\{0,1\}^{C}$ is its label vector over $C$ semantic categories. The same notation is used for both single-label and multi-label retrieval: $y_{ik}=1$ indicates that $\mathbf{x}_i$ is associated with the $k$-th category, and $y_{ik}=0$ otherwise.

Deep image hashing aims to learn a mapping $\mathcal{H}(\cdot)$ that encodes each image into a compact binary representation,
\begin{equation}
    \mathcal{H}:\mathbf{x}_i\mapsto \mathbf{b}_i\in\{-1,+1\}^{B},
\end{equation}
where $B$ is the code length. In most deep hashing frameworks, a continuous feature $\mathbf{f}_i\in\mathbb{R}^{B}$ is first produced by a neural network, and the binary code is then obtained by element-wise binarization
\begin{equation}
    \mathbf{b}_i = \operatorname{sgn}(\mathbf{f}_i).
\end{equation}
Retrieval is performed in the Hamming space induced by the binary codes.

For supervised image retrieval, semantic neighborhood relationships are derived from label overlap. Specifically, the pairwise semantic affinity between $\mathbf{x}_i$ and $\mathbf{x}_j$ is defined as
\begin{equation}
    s_{ij}=\mathbb{I}(\mathbf{y}_i^{\top}\mathbf{y}_j>0),
\end{equation}
where $s_{ij}=1$ indicates that the two images share at least one semantic label, and $s_{ij}=0$ otherwise. The objective is therefore to preserve such semantic affinities in the learned Hamming space, so that semantically related images are assigned short Hamming distances while unrelated images are separated.

A central limitation of the prevalent post-quantization paradigm is that representation learning and code formation are largely decoupled: the backbone is optimized in a continuous feature space, whereas retrieval is eventually conducted after a non-differentiable sign operation. This feature-to-code discrepancy becomes particularly undesirable when the binary code is expected to serve as the final retrieval representation. To mitigate this mismatch, we propose to learn hash-oriented representations natively within a Vision Transformer, rather than appending a terminal hashing layer after continuous feature extraction.

\subsection{Overview of HashViT}
\label{subsec:overview}

Fig.~\ref{framework} illustrates the overall architecture of the proposed HashViT. Given an input image $\mathbf{x}_i$, we divide it into $P$ non-overlapping patches and project them into $d$-dimensional patch tokens. In addition to the standard class token, we insert a dedicated hash token into the input sequence:
\begin{equation}
    \mathbf{Z}_0=    [\mathbf{z}_{\mathrm{cls}}^{0},\mathbf{t}^{0},\mathbf{z}_{1}^{0},\ldots,\mathbf{z}_{P}^{0}]
    +\mathbf{E}_{\mathrm{pos}},
\end{equation}
where $\mathbf{Z}_0\in\mathbb{R}^{(P+2)\times d}$, $\mathbf{z}_{\mathrm{cls}}^{0}$ is the class token, $\mathbf{t}^{0}$ is the hash token, $\{\mathbf{z}_{p}^{0}\}_{p=1}^{P}$ are patch tokens, and $\mathbf{E}_{\mathrm{pos}}$ denotes the positional embedding.

HashViT follows a dual-token design. The class token maintains a continuous global representation for semantic modeling, whereas the hash token is assigned to carry the binary-oriented retrieval representation. After each transformer block, the hash token is refined by a lightweight Hash Refinement Adapter (HRA). The first $B$ dimensions of the hash token are treated as a \emph{Hash Register}, from which the final binary code is generated, while the remaining $d-B$ dimensions form a \emph{Semantic Workspace} that supplies auxiliary continuous semantics for register refinement.

Formally, for the $l$-th transformer block, $l\in\{1,\ldots,L\}$, we first obtain
\begin{equation}
    \widehat{\mathbf{Z}}_l = \mathcal{F}_l(\mathbf{Z}_{l-1}),
\end{equation}
where $\mathcal{F}_l(\cdot)$ denotes the $l$-th ViT block and $\widehat{\mathbf{Z}}_l$ is the token sequence before hash refinement. Let $\widehat{\mathbf{t}}_l\in\mathbb{R}^{d}$ be the corresponding hash token. HRA produces the refined hash token as
\begin{equation}
    \mathbf{t}_l = \operatorname{HRA}(\widehat{\mathbf{t}}_l),
\end{equation}
which is then reinserted into the token sequence and passed to the next transformer block. The same HRA parameters are shared across transformer layers, which keeps the refinement module lightweight and enforces a consistent workspace-to-register correction. HRA only updates the Hash Register and keeps the Semantic Workspace unchanged, as detailed in Subsection~\ref{subsec:hra}.

Through this layer-wise refinement, the formation of the binary-oriented state is carried inside the transformer: the hash token interacts with visual tokens throughout the network and progressively evolves toward a binary-friendly state, while the final binarization remains a single sign operation at inference.

The model is trained with three complementary objectives. A learnable semantic hash center loss provides class-level anchors in the hash space, a similarity distillation loss transfers instance-level relational structure from the continuous class-token representation to the hash representation, and a quantization loss encourages each register dimension to approach the binary vertices $\{-1,+1\}$. During inference, only the final Hash Register is binarized and used for Hamming-distance retrieval.

\subsection{HASH Token for Native Hash Representation Learning}
\label{subsec:hash_token}

Existing deep hashing methods commonly attach a hashing projection to the output feature of a CNN or Transformer backbone. 
Although convenient, such a design treats binary code generation as a post-processing step, so the backbone is not explicitly encouraged to organize its intermediate representations toward the discrete Hamming space used at retrieval.

A straightforward alternative is to impose quantization constraints on intermediate hidden features. However, this strategy is overly restrictive because the same hidden representation must simultaneously support continuous semantic abstraction and binary code generation. Such a constraint can impair the expressive capacity of visual tokens, especially in a transformer where token interactions rely on continuous feature geometry.

HashViT addresses this issue by introducing a dedicated hash token. Unlike the class token, whose role is to summarize continuous image semantics, the hash token is responsible for learning a retrieval-oriented representation. Since it participates in self-attention from the first layer, the hash token can aggregate visual evidence and exchange information with patch tokens during the entire forward pass. Binary code learning is thus integrated into the backbone, rather than isolated in an external projection head.

Given the intermediate hash token $\widehat{\mathbf{t}}_l$ produced by the $l$-th transformer block, we decompose it into two subspaces:
\begin{equation}
\label{eq:decompose}
    \widehat{\mathbf{t}}_l=[\widehat{\mathbf{u}}_l;\widehat{\mathbf{v}}_l],
\end{equation}
where $\widehat{\mathbf{u}}_l\in\mathbb{R}^{B}$ is the Hash Register, $\widehat{\mathbf{v}}_l\in\mathbb{R}^{d-B}$ is the Semantic Workspace, and $[\cdot;\cdot]$ denotes concatenation. The Hash Register is explicitly aligned with the target Hamming space, whereas the Semantic Workspace retains complementary continuous semantics that are useful for refining the register.

After the final transformer layer, the binary code of $\mathbf{x}_i$ is generated solely from the final Hash Register $\mathbf{u}_{i,L}$:
\begin{equation}
\label{eq:final_code}
\mathbf{h}_i=\tanh(\mathbf{u}_{i,L}),~
    \mathbf{b}_i=\operatorname{sgn}(\mathbf{h}_i)\in\{-1,+1\}^{B}.
\end{equation}
The remaining dimensions of the hash token, together with the class and patch tokens, are discarded for retrieval.

This structured token design has two important effects. First, it confines binary constraints to the retrieval-specific subspace, allowing the class and patch tokens to preserve continuous semantic modeling ability. 
Second, it narrows the gap between continuous training representations and discrete retrieval codes by letting the Hash Register be repeatedly updated inside the transformer before the final sign operation, as quantified by the layer-wise quantization error in Sec.~\ref{layer}.
This design also avoids repeatedly projecting a high-dimensional continuous token into the hash space, keeping the hash state as an explicit part of the token representation.
\subsection{Hash Refinement Adapter}
\label{subsec:hra}

Inspired by the observation that additional register-like representations can provide useful workspace for Vision Transformers~\cite{vitregisters}, we split the HASH token into two parts: a low-dimensional Hash Register for code generation and a higher-dimensional Semantic Workspace for auxiliary semantic modeling. The Hash Register is directly used to produce binary codes, while the Semantic Workspace is not binarized. Therefore, the role of the Hash Refinement Adapter (HRA) is simple: it uses the workspace to provide a semantic correction to the register.

Recall from Eq.~\eqref{eq:decompose} that the HASH token after the $l$-th transformer block is decomposed as $\widehat{\mathbf{t}}_l=[\widehat{\mathbf{u}}_l;\widehat{\mathbf{v}}_l]$, with Hash Register $\widehat{\mathbf{u}}_l\in\mathbb{R}^{B}$ and Semantic Workspace $\widehat{\mathbf{v}}_l\in\mathbb{R}^{d-B}$. HRA first maps the workspace to the register dimension:

\begin{equation}
    \mathbf{u}_l
    =
    \widehat{\mathbf{u}}_l + \operatorname{MLP}_{\mathrm{HRA}}(\widehat{\mathbf{v}}_l),
\end{equation}
where $\operatorname{MLP}_{\mathrm{HRA}}:\mathbb{R}^{d-B}\rightarrow\mathbb{R}^{B}$ is a lightweight shared mapping applied after each transformer block and is implemented as a single linear layer. The Hash Register is thus updated by a residual correction from the workspace. The refined HASH token is reconstructed as

\begin{equation}
    \mathbf{t}_l =
    [\mathbf{u}_l;\widehat{\mathbf{v}}_l],
\end{equation}
and is used as the HASH token input to the next transformer block.

This design keeps HRA deliberately lightweight. It does not transform all tokens or the whole HASH token. Instead, it only updates the binary-oriented register using information from the semantic workspace. 

The residual form preserves the register's original evolution while injecting complementary semantics from the workspace, providing a simple bridge between continuous semantic modeling and binary code learning.

\subsection{Semantic Supervision for Native Hash Learning}
\label{subsec:supervision}
Before introducing the supervision terms, we define the bounded continuous hash representation as $\mathbf{h}_i=\tanh(\mathbf{u}_{i,L})$, following Sec.~\ref{subsec:hash_token}. The hash-side representation involved in the supervision losses is computed on $\mathbf{h}_i$ before the final sign operation. Since $\tanh(\cdot)$ is monotonic, it bounds the training representation without changing the final binary code.

\paragraph{Learnable semantic hash center loss.}
We maintain a set of class-wise learnable semantic centers in the hash space
$\mathcal{C}=\{\mathbf{c}_k\}_{k=1}^{C}$, where
$\mathbf{c}_k\in\mathbb{R}^{B}$ denotes the learnable semantic center of class $k$ in the hash space.
Following the idea of center or proxy-based supervision in metric learning and deep hashing~\cite{CenterHash, movshovitz2017no, kim2020proxy, CSQ}, we use these centers to provide global semantic targets for hash representation learning.

To introduce semantic priors, each center is initialized from a text-guided class prototype. Let $\mathbf{e}_k=E_t(\pi_k)\in\mathbb{R}^{512}$ denote the text embedding of class $k$, where $\pi_k$ is the class-name prompt and $E_t(\cdot)$ is the frozen CLIP text encoder~\cite{clip}. The initialization function $g(\cdot)$ maps the $512$-dimensional text embedding to the target $B$-dimensional hash space via a fixed random orthogonal projection followed by $\ell_2$ normalization:
\begin{equation}
    \mathbf{c}_k^{0}
    =
    g(\mathbf{e}_k)
    =
    \frac{\mathbf{R}_B^{\top}\mathbf{e}_k}
    {\|\mathbf{R}_B^{\top}\mathbf{e}_k\|_2+\epsilon}.
\end{equation}

The random orthogonal projection provides a parameter-free dimensionality reduction that approximately preserves the angular structure of text prototypes in a low-dimensional space. It is used only for center initialization and introduces no trainable parameters. After initialization, the centers are optimized as learnable parameters, while the CLIP text encoder and the projection matrix are not used during inference.

For compact notation, we define the normalized image-to-center similarity as
\begin{equation}
    \rho_{ik}=
\left\langle
\frac{\mathbf{h}_i}{\|\mathbf{h}_i\|_2+\epsilon},
\frac{\mathbf{c}_k}{\|\mathbf{c}_k\|_2+\epsilon}
\right\rangle .
\end{equation}

Here, $\epsilon$ is a small constant used to avoid division by zero. For exclusive single-label retrieval, each sample has one positive center and all remaining centers can be regarded as negatives. 
Let $\mathcal{I}$ denote the mini-batch index set with $|\mathcal{I}|=n$.
For each class $k$, we define
$\mathcal{P}_k=\{i\in\mathcal{I}\mid y_{ik}=1\}$,
$\mathcal{N}_k=\{i\in\mathcal{I}\mid y_{ik}=0\}$, and
$\mathcal{K}^{+}=\{k\mid |\mathcal{P}_k|>0\}$.

The single-label center loss is defined as
\begin{equation}
\begin{aligned}
\mathcal{L}_{\mathrm{single}}
=&
\frac{1}{|\mathcal{K}^{+}|+\epsilon}
\sum_{k\in\mathcal{K}^{+}}
\log\left(
1+\sum_{i\in\mathcal{P}_k}
e^{-\alpha(\rho_{ik}-\delta)}
\right) \\
&+
\frac{1}{C}
\sum_{k=1}^{C}
\log\left(
1+\sum_{i\in\mathcal{N}_k}
e^{\alpha(\rho_{ik}+\delta)}
\right),
\end{aligned}
\end{equation}
where $\delta$ is the margin and $\alpha$ is the scale factor.
The first term pulls samples toward their positive centers, while the second term pushes them away from negative centers.

For non-exclusive multi-label retrieval, an image may correspond to multiple semantic centers. Instead of independently pulling the image toward every positive center, we maximize the probability mass assigned to its positive center set. Let
$\mathcal{Y}_i^{+}=\{k\mid y_{ik}=1\}$.
The multi-label center loss is
\begin{equation}
\mathcal{L}_{\mathrm{multi}}
=
-\frac{1}{n}
\sum_{i=1}^{n}
\log
\frac{
\sum_{k\in\mathcal{Y}_i^{+}} e^{\gamma \rho_{ik}}
}{
\sum_{k=1}^{C} e^{\gamma \rho_{ik}}
},
\end{equation}
where $\gamma$ is a scale factor. This positive-set formulation is suitable for multi-label data because it encourages the hash representation to be close to the relevant semantic set as a whole.

Finally, the semantic hash center loss is selected according to the label structure:
\begin{equation}
\mathcal{L}_{c}=
\begin{cases}
\mathcal{L}_{\mathrm{single}}, & \text{for exclusive single-label data},\\
\mathcal{L}_{\mathrm{multi}}, & \text{for non-exclusive multi-label data}.
\end{cases}
\end{equation}

\paragraph{Similarity distillation loss.}
Class centers provide category-level supervision, but they do not explicitly preserve fine-grained instance relationships within a mini-batch. To complement center learning, we distill the relational structure of continuous class-token features into the hash representations.
Let $\mathbf{g}_i\in\mathbb{R}^{d}$ denote the final class-token representation of image $\mathbf{x}_i$. The teacher feature is detached when computing this objective. We normalize the teacher and hash representations as
\begin{equation}
    \overline{\mathbf{g}}_i=\frac{\operatorname{sg}(\mathbf{g}_i)}{\|\operatorname{sg}(\mathbf{g}_i)\|_2+\epsilon},\quad
    \overline{\mathbf{h}}_i=\frac{\mathbf{h}_i}{\|\mathbf{h}_i\|_2+\epsilon},
\end{equation}
where $\operatorname{sg}(\cdot)$ denotes stop-gradient. The teacher-side and hash-side similarity matrices are computed by
\begin{equation}
    S^{t}_{ij}=\overline{\mathbf{g}}_i^{\top}\overline{\mathbf{g}}_j,
    \quad
    S^{h}_{ij}=\overline{\mathbf{h}}_i^{\top}\overline{\mathbf{h}}_j.
\end{equation}

The similarity distillation loss is the mean squared discrepancy between the two matrices:
\begin{equation}
    \mathcal{L}_{sd}
    =\frac{1}{n^2}\sum_{i=1}^{n}\sum_{j=1}^{n}
    \left(S^{h}_{ij}-S^{t}_{ij}\right)^2.
\end{equation}
This objective encourages the Hash Register to preserve the pairwise topology encoded by the continuous class-token representation, while the stop-gradient operation prevents the hash objective from altering the teacher feature space.

\paragraph{Quantization loss.}
Although the HASH token is designed to learn binary-oriented representations natively, we optionally impose a lightweight quantization regularizer on the bounded hash representation:
\begin{equation}
    \mathcal{L}_{q}
    =
    \frac{1}{nB}\sum_{i=1}^{n}\sum_{b=1}^{B}
    (|h_{ib}|-1)^2,
\end{equation}
where $h_{ib}$ is the $b$-th element of $\mathbf{h}_i$. This loss encourages each dimension to approach the binary vertices $\{-1,+1\}$ and reduces the discrepancy caused by the final sign operation. It is applied only to the Hash Register and does not constrain the class token, patch tokens, or Semantic Workspace. As shown in the experiments, this term serves mainly as an auxiliary stabilizer rather than the primary source of performance gain.

\setlength{\heavyrulewidth}{0.9pt}
\setlength{\lightrulewidth}{0.5pt}
\setlength{\cmidrulewidth}{0.5pt}
\renewcommand{\arraystretch}{1.35}
\setlength{\tabcolsep}{4pt}
\begin{table*}[!htbp]
\caption{Performance comparison of deep hashing methods.}
\centering
\resizebox{\textwidth}{!}{
\begin{tabular}{@{}l|l|cccc|cccc|cccc@{}}
\toprule
\multirow{3}{*}[-6pt]{\bfseries Methods} & 
\multirow{3}{*}[-6pt]{\bfseries Reference} & 
\multicolumn{12}{c@{}}{\bfseries Datasets (\%)} \\
\cmidrule(lr){3-14}
& & 
\multicolumn{4}{c|}{\bfseries CIFAR-10@54000} & 
\multicolumn{4}{c|}{\bfseries NUS-WIDE@5000} & 
\multicolumn{4}{c@{}}{\bfseries IMAGENET@1000} \\
\cmidrule(lr){3-6} \cmidrule(lr){7-10} \cmidrule(lr){11-14}
& & 
16bit & 32bit & 48bit & 64bit & 
16bit & 32bit & 48bit & 64bit & 
16bit & 32bit & 48bit & 64bit \\
\midrule
DPSH \cite{DPSH} & IJCAI16 
& 72.30 & 74.70 & 75.70 & 75.50
& 70.46 & 72.34 & 73.40 & 73.80
& 45.66 & 44.59 & 48.54 & 53.60 \\
HashNet \cite{HashNet} & ICCV17 
& 73.21 & 76.32 & 78.20 & 79.12
& 66.12 & 69.32 & 70.88 & 72.32
& 53.85 & 60.12 & 64.55 & 67.14 \\
DCH \cite{DCH} & CVPR18 
& 76.56 & 78.33 & 77.98 & 77.33
& 67.54 & 68.47 & 69.38 & 67.98
& 53.63 & 57.52 & 55.85 & 54.71 \\
DPN \cite{DPN} & IJCAI20
& 75.18 & 76.92 & 76.24 & 76.28
& 66.12 & 69.48 & 71.86 & 71.89
& 59.82 & 66.65 & 68.71 & 69.36 \\
CSQ \cite{CSQ} & CVPR20 
& 80.67 & 81.98 & 83.17 & 84.63
& 72.91 & 74.17 & 74.36 & 75.02
& 85.10 & 86.50 & 86.90 & 87.30 \\
TransHash* \cite{transhash} & ICMR22 
& 90.75 & 91.08 & 91.41 & 91.66
& 72.63 & 73.93 & 75.32 & 74.88
& 78.52 & 87.33 & 89.32 & 89.21 \\
CenterHash \cite{CenterHash} & CVPR23 
& 83.25 & 84.26 & 84.55 & 84.93
& 73.78 & 78.43 & 79.85 & 80.56
& 84.40 & 87.36 & 88.03 & 88.52 \\
MSViT-S* \cite{msvit} & TMM24 
& 84.29 & 88.67 & 89.38 & 90.00
& - & - & - & -
& 72.34 & 82.99 & 86.53 & 87.91 \\
FTH-DeiT-S \cite{FTH} & ACMMM25 
& 90.43 & 90.24 & 91.81 & 91.93
& 75.45 & 77.66 & 77.90 & 77.83
& 81.60 & 84.19 & 85.16 & 85.20 \\
SHC \cite{SHC} & TOIS25 
& 81.43 & 84.37 & 82.85 & 83.14
& 70.11 & 71.93 & 71.74 & 72.06
& 77.80 & 78.85 & 79.09 & 78.83 \\
NDBE \cite{NDBE} & TMM26 
& 82.64 & 84.89 & 84.07 & 85.41
& 70.25 & 72.95 & 74.04 & 75.04 
& 38.45 & 47.30 & 51.22 & 53.89 \\
CRH \cite{CRH} & AAAI26 
& 79.37 & 81.51 & 84.23 & 81.89
& 72.77 & 75.57 & 76.29 & 76.79
& 74.69 & 78.33 & 78.65 & 79.22 \\
\midrule
HashViT-T & (Ours) 
& 84.94 & 87.95 & 88.73 & 89.05
& 72.67 & 76.14 & 77.38 & 77.63
& 73.12 & 74.77 & 82.90 & 83.52 \\
\textbf{HashViT-S} & \textbf{(Ours)} 
& \textbf{92.67} & \textbf{93.59} & \textbf{95.68} & \textbf{96.36}
& \textbf{75.61}& \textbf{78.96} & \textbf{80.16} & \textbf{81.02}
& \textbf{86.85} & \textbf{90.12} & \textbf{90.68} & \textbf{91.06} \\
\bottomrule
\end{tabular}
}
\label{tab:hash_comparison}
\end{table*}

\renewcommand{\arraystretch}{1}

\subsection{Overall Objective and Retrieval Inference}
\label{subsec:objective_inference}

The overall training objective of HashViT is
\begin{equation}
    \mathcal{L}
    =
    \mathcal{L}_{c}
    +
    \lambda_{sd}\mathcal{L}_{sd}
    +
    \lambda_{q}\mathcal{L}_{q},
\end{equation}
where $\lambda_{sd}$ and $\lambda_q$ control the weights of similarity distillation and quantization regularization, respectively. The semantic center loss $\mathcal{L}_{c}$ provides class-level supervision, $\mathcal{L}_{sd}$ preserves instance-level relational structure, and $\mathcal{L}_{q}$ stabilizes the final binarization.

During training, the hash-side representation used in the above losses is the bounded continuous representation $\mathbf{h}_i=\tanh(\mathbf{u}_{i,L})$, where $\mathbf{u}_{i,L}$ is the final Hash Register. During inference, the binary code is obtained as in Eq.~\eqref{eq:final_code}, and only the final Hash Register is used for retrieval, while the class token, patch tokens, and Semantic Workspace are discarded.

Given a query image $\mathbf{x}_q$, HashViT produces its binary code $\mathbf{b}_q$. The binary codes of database images are precomputed and stored. Retrieval is performed by ranking database images according to the Hamming distance
\begin{equation}
    d_H(\mathbf{b}_q,\mathbf{b}_i)
    =
    \frac{1}{2}
    \left(
    B-\mathbf{b}_q^{\top}\mathbf{b}_i
    \right),
\end{equation}
where a smaller distance indicates higher retrieval priority.

The pretrained text encoder and the random orthogonal projection are used only for semantic center initialization and are not involved in model forward passes after center initialization or inference. Therefore, HashViT preserves the standard efficiency advantage of deep hashing: each image is represented by a compact $B$-bit code, and retrieval only requires efficient Hamming-distance comparison.

\section{Experiment}\label{exp}
\subsection{Baselines and Datasets}
We compare the proposed HashViT with a broad range of representative deep hashing methods on three widely used benchmarks: CIFAR‑10, NUS‑WIDE, and IMAGENET, including DPSH~\cite{DPSH}, HashNet~\cite{HashNet}, DCH~\cite{DCH}, DPN~\cite{DPN}, CSQ~\cite{CSQ}, TransHash~\cite{transhash}, CenterHash~\cite{CenterHash}, MSViT-S~\cite{msvit}, FTH-DeiT-S~\cite{FTH}, SHC~\cite{SHC}, NDBE~\cite{NDBE}, and CRH~\cite{CRH}. The methods marked with an asterisk (*) denote results reported in the original papers. Results without * are reproduced using the official implementations or our re-implementation under the same data split and evaluation protocol.

\textbf{CIFAR-10} is a single‑label dataset comprising 60,000 32$\times$32 color images evenly distributed over 10 classes. We follow the same setup as the experiment in~\cite{transhash,MambaHash}, and randomly sample 5,000 images (500 per class) for training, 1,000 images (100 per class) as queries, and treat the remaining 54,000 images as the retrieval database.

\textbf{NUS-WIDE} is a multi-label web image dataset containing 269,648 images annotated with 81 concept labels. Following the protocol in~\cite{transhash,MambaHash}, two images are regarded as semantically relevant if they share at least one label. We randomly sample 5,000 images as queries and use the remaining images as the retrieval database. From the database, 10,000 images are further sampled for training.

\textbf{IMAGENET} is the standard benchmark from the Large‑Scale Visual Recognition Challenge (ILSVRC 2015). We follow the experimental setup in~\cite{transhash,MambaHash} by randomly choosing 100 classes. The full training set of these 100 classes (128,503 images) serves as the retrieval database, while 5,000 validation images (50 per class) are used as queries. For training, we randomly sample 130 images per class, yielding 13,000 training images in total.

\subsection{Implementation Details and Metrics}
\subsubsection{Implementation Details.}
All input images are first resized to $256\times256$ and then randomly cropped to $224\times224$ with random horizontal flipping for data augmentation. HashViT is implemented in PyTorch~\cite{pytorch}. All models are trained for 100 epochs with a batch size of 512. We use Adam~\cite{adam} as the optimizer, with the learning rate and weight decay both set to $1\times10^{-5}$. All experiments are conducted on a workstation equipped with an Intel Xeon Silver 4214 CPU, 125 GB RAM, and an NVIDIA Tesla A800 GPU. The source code and additional implementation details will be made publicly available.

For the loss weights, we use dataset-specific settings selected according to the parameter sensitivity analysis. Specifically, $(\lambda_{sd},\lambda_q)$ is set to $(1,0)$ on CIFAR-10, $(1,0)$ on NUS-WIDE, and $(10,1)$ on IMAGENET. For the semantic hash center loss, the scale and margin parameters in the single-label formulation are fixed as $\alpha=32$ and $\delta=0.1$, following the common setting of proxy-based metric learning losses~\cite{kim2020proxy,movshovitz2017no}. The scale factor in the multi-label positive-set formulation is fixed as $\gamma=24$, following the use of scaled cosine similarities in center-based hashing objectives~\cite{CSQ,SHC}. These parameters are kept fixed across the corresponding experiments without dataset-specific tuning.

\subsubsection{Evaluation Metrics.}
We evaluate retrieval performance using mean Average Precision (mAP) under different hash code lengths, including 16, 32, 48, and 64 bits. Following common practice in related work~\cite{transhash,HybridHash,MambaHash}, mAP is computed based on the top 54,000 returned samples for CIFAR-10, the top 5,000 returned samples for NUS-WIDE, and the top 1,000 returned samples for IMAGENET. In addition to mAP, we also report Precision-Recall (PR) curves and Precision@Top-$K$ curves to further analyze retrieval behavior. PR curves evaluate the precision-recall trade-off under different retrieval thresholds, while Precision@Top-$K$ measures the proportion of relevant samples among the top-$K$ retrieved results.

\setlength{\heavyrulewidth}{0.9pt}
\setlength{\lightrulewidth}{0.5pt}
\setlength{\cmidrulewidth}{0.5pt}
\renewcommand{\arraystretch}{1.45}
\setlength{\tabcolsep}{4pt}

\begin{table*}[!htbp]
\caption{Backbone replacement analysis with HashViT-S. "Method+HashViT-S" denotes replacing the original visual backbone with HashViT-S while keeping the corresponding hashing objective.}
\centering
\resizebox{\textwidth}{!}{
\begin{tabular}{@{}l|cccc|cccc|cccc@{}}
\toprule
\multirow{3}{*}[-6pt]{\bfseries Methods} & 
\multicolumn{12}{c@{}}{\bfseries Datasets (\%)} \\
\cmidrule(lr){2-13}
& \multicolumn{4}{c|}{\bfseries CIFAR-10@54000} 
& \multicolumn{4}{c|}{\bfseries NUS-WIDE@5000} 
& \multicolumn{4}{c@{}}{\bfseries IMAGENET@1000} \\
\cmidrule(lr){2-5} \cmidrule(lr){6-9} \cmidrule(lr){10-13}
& 16bit & 32bit & 48bit & 64bit 
& 16bit & 32bit & 48bit & 64bit 
& 16bit & 32bit & 48bit & 64bit \\
\midrule
DPSH \cite{DPSH}
& 72.30 & 74.70 & 75.70 & 75.50
& 70.46 & 72.34 & 73.40 & 73.80
& 45.66 & 44.59 & 48.54 & 53.60 \\
DPSH+HashViT-S
& \textbf{78.96} & \textbf{78.91} & \textbf{79.95} & \textbf{78.79}
& \textbf{74.83} & \textbf{75.83} & \textbf{76.17} & \textbf{76.44}
& \textbf{74.59} & \textbf{77.31} & \textbf{79.41} & \textbf{78.73} \\
HashNet \cite{HashNet}
& 73.21 & 76.32 & 78.20 & 79.12
& 66.12 & 69.32 & 70.88 & 72.32
& 53.85 & 60.12 & 64.55 & 67.14 \\
HashNet+HashViT-S
& \textbf{89.18} & \textbf{95.28} & \textbf{95.74} & \textbf{95.27}
& \textbf{73.53} & \textbf{74.58} & \textbf{75.89} & \textbf{75.88}
& \textbf{78.20} & \textbf{84.89} & \textbf{85.12} & \textbf{86.01} \\
DCH \cite{DCH}
& 76.56 & 78.33 & 77.98 & 77.33
& 67.54 & 68.47 & 69.38 & 67.98
& 53.63 & 57.52 & 55.85 & 54.71 \\
DCH+HashViT-S
& \textbf{91.75} & \textbf{92.57} & \textbf{92.48} & \textbf{93.74}
& \textbf{74.06} & \textbf{73.50} & \textbf{72.61} & \textbf{74.43}
& \textbf{87.34} & \textbf{86.33} & \textbf{85.16} & \textbf{86.37} \\
DPN \cite{DPN}
& 75.18 & 76.92 & 76.24 & 76.28
& 66.12 & 69.48 & 71.86 & 71.89
& 59.82 & 66.65 & 68.71 & 69.36 \\
DPN+HashViT-S
& \textbf{94.84} & \textbf{93.74} & \textbf{95.09} & \textbf{93.73}
& \textbf{72.72} & \textbf{74.54} & \textbf{75.53} & \textbf{75.91}
& \textbf{87.78} & \textbf{89.59} & \textbf{89.25} & \textbf{89.15} \\
CSQ \cite{CSQ}
& 80.67 & 81.98 & 83.17 & 84.63
& 72.91 & 74.17 & 74.36 & 75.02
& 85.10 & 86.50 & 86.90 & 87.30 \\
CSQ+HashViT-S
& \textbf{93.63} & \textbf{92.16} & \textbf{90.26} & \textbf{93.95}
& \textbf{72.94} & \textbf{74.32} & \textbf{74.45} & \textbf{75.96}
& \textbf{86.83} & \textbf{89.59} & \textbf{89.91} & \textbf{89.43} \\
CenterHash \cite{CenterHash}
& 83.25 & 84.26 & 84.55 & 84.93
& 73.78 & 78.43 & 79.85 & 80.56
& 84.40 & 87.36 & 88.03 & 88.52 \\
CenterHash+HashViT-S
& \textbf{92.92} & \textbf{93.46} & \textbf{94.06} & \textbf{94.73}
& \textbf{74.19} & \textbf{78.55} & \textbf{79.91} & \textbf{80.73}
& \textbf{87.04} & \textbf{89.95} & \textbf{89.56} & \textbf{89.69} \\
\bottomrule
\end{tabular}
}
\label{tab:backbone_comparison}
\end{table*}
\renewcommand{\arraystretch}{1}

\subsection{Quantitative Comparison}

\subsubsection{Comparison with Representative Hashing Methods}
Table~\ref{tab:hash_comparison} reports the mAP comparison between HashViT and representative deep hashing methods on CIFAR-10, NUS-WIDE, and IMAGENET. The compared methods cover pairwise-supervised hashing methods, such as DPSH~\cite{DPSH}, HashNet~\cite{HashNet}, DCH~\cite{DCH}, and DPN~\cite{DPN}; center-based hashing methods, such as CSQ~\cite{CSQ}, CenterHash~\cite{CenterHash}, and SHC~\cite{SHC}; and transformer-based hashing methods, such as TransHash~\cite{transhash}, MSViT-S~\cite{msvit}, and FTH-DeiT-S~\cite{FTH}. The methods marked with an asterisk (*) denote results reported in the original papers.

Overall, HashViT-S achieves the best or comparable performance across the three datasets, with the largest margins on CIFAR-10 and ImageNet and competitive results on the multi-label NUS-WIDE.
Compared with conventional pairwise hashing methods, the advantage mainly comes from replacing local pairwise supervision with more stable semantic hash space organization. Compared with center-based methods such as CSQ, CenterHash, and SHC, HashViT further improves retrieval accuracy by introducing learnable text-guided semantic centers, which provide more adaptive and semantically meaningful supervision for hash representation learning. More importantly, compared with transformer-based baselines, HashViT does not rely on a post-quantization hash projection layer. 
Instead, it learns the hash representation inside the transformer through a dedicated HASH token, moving the formation of the binary-oriented representation into the backbone, while the final binarization remains a single sign operation at inference.

The results also show a favorable accuracy-efficiency trade-off. HashViT-T uses only 5.54M parameters but maintains competitive retrieval performance, while HashViT-S achieves stronger results with a compact ViT-Small backbone. In particular, HashViT-S surpasses large transformer-based hashing methods such as TransHash, although these methods use much larger ViT-Large backbones. 
This suggests that the gain is not solely due to model capacity, but also reflects the proposed design—native in-transformer hash representation learning together with learnable semantic hash centers—although Table~\ref{tab:backbone_comparison} indicates that the strong ViT backbone itself accounts for a substantial portion of the improvement.

\subsubsection{Comparison under HashViT-S Backbone}
Table~\ref{tab:backbone_comparison} further studies whether existing hashing objectives can benefit from the HashViT-S backbone. Specifically, several representative hashing methods are re-implemented by replacing their original visual backbone with HashViT-S while keeping their original hashing objectives. This comparison helps separate the contribution of the backbone representation from the contribution of the proposed native hash learning objective.

The results indicate that the HashViT-S architecture can substantially improve several existing hashing objectives, suggesting that the proposed token-level representation design is broadly useful. However, the relative performance varies across objectives and code lengths. This implies that the backbone design and the training objective are complementary: some conventional losses can benefit from the proposed backbone, while the complete HashViT provides the most consistent overall behavior across datasets, especially in the 64-bit setting.

\subsection{Ablation Study}\label{Ablation}
We conduct ablation studies to examine the key components of HashViT, including the dedicated HASH token, the Hash Refinement Adapter (HRA), the training objectives, and the semantic hash center initialization. Unless otherwise specified, all variants are evaluated under the same training protocol, backbone scale, and hash code length unless otherwise specified. The results are reported in terms of mAP on CIFAR-10, NUS-WIDE, and IMAGENET.

\subsubsection{Effect of the Dedicated HASH Token}

We first verify whether native hash learning benefits from a dedicated hash-oriented token. As shown in Table~\ref{tab:ablation_token_design}, HashViT-S consistently outperforms FTH-DeiT-S and HashDeiT-S. FTH-DeiT-S follows the conventional post-quantization hashing paradigm, where binary codes are generated by a terminal hash layer after continuous feature extraction. In contrast, HashViT-S introduces the HASH token into the transformer and learns binary-oriented representations during feature modeling. 
The clear improvement over FTH-DeiT-S indicates that the gain does not merely come from adopting a DeiT/ViT-style backbone, but from moving hash representation learning into the transformer.

HashViT-S also performs better than HashDeiT-S, which reuses the DeiT distillation token as the hash carrier. This comparison further shows that an existing learnable token is not naturally suitable for hash learning. The distillation token in DeiT is originally designed to absorb teacher-guided semantic supervision, while the proposed HASH token is explicitly assigned to binary-oriented representation learning for retrieval. Therefore, the superior performance of HashViT-S comes from the specialized role of the HASH token, rather than from simply adding or reusing a learnable token.

\begin{table}[t]
\centering
\caption{Ablation study on different learnable token methods with 64-bit.}
\label{tab:ablation_token_design}
\setlength{\tabcolsep}{8pt}
\begin{tabular}{@{}l|ccc@{}}
\toprule
\bfseries Method & \bfseries CIFAR-10 & \bfseries NUS-WIDE & \bfseries IMAGENET \\
\midrule
FTH-DeiT-S \cite{FTH} & 91.93 & 77.83 & 85.20 \\
HashDeiT-S & 93.54 & 79.12 & 90.12 \\
HashViT-S & \textbf{96.36} & \textbf{81.02} & \textbf{91.06} \\
\bottomrule
\end{tabular}
\end{table}
\begin{table}[t]
\centering
\caption{Ablation study on the Hash Refinement Adapter with 64-bit.}
\label{tab:ablation_hra}
\renewcommand{\arraystretch}{1.25}
\setlength{\tabcolsep}{8pt}
\begin{tabular}{@{}l|ccc@{}}
\toprule
\bfseries Method & \bfseries CIFAR-10 & \bfseries NUS-WIDE & \bfseries IMAGENET \\
\midrule
w/o HRA & 95.88 & 80.31 & 90.47 \\
HashViT-S & \textbf{96.36} & \textbf{81.02} & \textbf{91.06} \\
\bottomrule
\end{tabular}
\end{table}

\subsubsection{Effect of Hash Refinement Adapter}

We then evaluate the contribution of HRA. As reported in Table~\ref{tab:ablation_hra}, removing HRA leads to a performance drop compared with the full model. This suggests that directly using the HASH token output is suboptimal, even though the token itself is already designed for hash learning. HRA further refines the Hash Register by incorporating information from the Semantic Workspace, allowing the binary-oriented component to receive semantic compensation without forcing the entire HASH token to satisfy strict quantization constraints. 
This result confirms that HRA is not a generic MLP add-on, but a refinement module coupled with the structured HASH token design.
This indicates that HRA, despite being lightweight, provides a small but consistent gain across datasets (Table~\ref{tab:ablation_hra}), acting as a refinement module coupled with the structured HASH token design rather than a generic MLP add-on.

\begin{table}[t]
\centering
\caption{Ablation study on the loss function of HashViT with 64-bit.}
\label{tab:ablation_losses}
\renewcommand{\arraystretch}{1.25}
\setlength{\tabcolsep}{8pt}
\begin{tabular}{@{}lcc|ccc@{}}
\toprule
\bfseries $\mathcal{L}_{c}$ & \bfseries $\mathcal{L}_{sd}$ & \bfseries $\mathcal{L}_{q}$ & \bfseries CIFAR-10 & \bfseries NUS-WIDE & \bfseries IMAGENET \\
\midrule
$\checkmark$ &&& 95.59 & 80.43 & 90.21 \\
$\checkmark$ & $\checkmark$ && \textbf{96.36} & \textbf{81.02} & 91.01 \\
$\checkmark$ & $\checkmark$ & $\checkmark$& \textbf{96.36} & \textbf{81.02} & \textbf{91.06} \\
\bottomrule
\end{tabular}
\end{table}
\begin{table}[t]
\centering
\caption{Ablation study on different semantic hash center initialization strategies with 64-bit.}
\label{tab:ablation_center_init}
\renewcommand{\arraystretch}{1.25}
\setlength{\tabcolsep}{8pt}
\begin{tabular}{@{}l|ccc@{}}
\toprule
\bfseries Initialization & \bfseries CIFAR-10 & \bfseries NUS-WIDE & \bfseries IMAGENET \\
\midrule
Random & 94.20 & 77.98 & 87.01 \\
SHC-guided \cite{SHC} & 94.46 & 78.29 & 88.47 \\
Text-guided & \textbf{96.36} & \textbf{81.02} & \textbf{91.06} \\
\bottomrule
\end{tabular}
\end{table}
\subsubsection{Effect of Training Objectives}

Table~\ref{tab:ablation_losses} analyzes the effects of the similarity distillation loss $\mathcal{L}_{sd}$ and the quantization loss $\mathcal{L}_q$. Removing $\mathcal{L}_{sd}$ decreases retrieval performance, showing that instance-level similarity structures from CLS representations provide useful guidance for the HASH token. This supervision complements class-level semantic center learning by preserving the relative topology among samples in the continuous semantic space.

The effect of $\mathcal{L}_q$ is relatively mild. Removing it leads to almost unchanged performance on CIFAR-10 and NUS-WIDE, and only a slight decrease on IMAGENET. This result is consistent with the motivation of HashViT: the HASH token can already evolve toward binary-friendly representations through native in-transformer learning, without relying on a strong quantization penalty. Therefore, $\mathcal{L}_q$ is better understood as a lightweight stabilizing regularizer for final binarization, rather than the main driver of retrieval performance.

\subsubsection{Effect of Semantic Hash Center Initialization}
Finally, we investigate different initialization strategies for semantic hash centers. As shown in Table~\ref{tab:ablation_center_init}, random initialization provides only weak semantic organization, while SHC-guided initialization offers limited improvement. In contrast, text-guided initialization achieves the best performance on all three datasets. This demonstrates that language-derived semantic priors provide a more meaningful starting point for organizing semantic centers, and the learnable optimization further adapts these centers to the visual hash space.

Overall, the ablation results support the main design choices of HashViT. The dedicated HASH token is more effective than post-hoc projection or reusing an existing distillation token; HRA further improves the structured hash representation; similarity distillation provides useful relational guidance; and quantization regularization mainly stabilizes the final binarization. These results verify that the performance gain of HashViT comes primarily from the coherent combination of native hash learning, structured token refinement, and semantic center supervision.

\begin{figure*}[t]
\centering
\includegraphics[width=\linewidth]{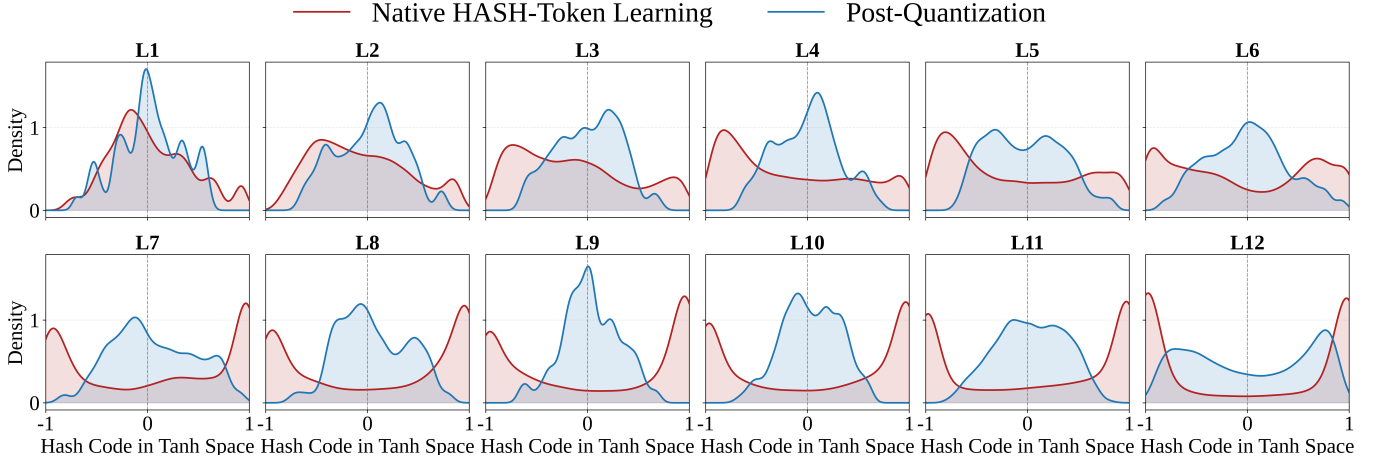}
\caption{Layer-wise Hash Code Distribution Dynamics under native Hash token learning and post-quantization. Native Hash token learning progressively drives hash-code distributions toward binary states across transformer layers, indicating that binary-friendly representations emerge through intrinsic transformer hash token evolution, whereas post-quantization relies on a final projection after feature extraction.}
\label{fig:layerwise_quantization}
\end{figure*}

\subsection{Layer-wise Quantization Analysis}
\label{layer}
To verify whether binary-oriented representations are formed inside the Transformer, we analyze the layer-wise evolution of the Hash Register. We compare HashViT with a controlled post-quantization baseline that uses the same ViT backbone and an additional learnable hash token, but generates binary codes only through a terminal hash projection. Since this baseline does not use the register-workspace decomposition or HRA, its hash token remains a continuous representation before the final projection.

For quantitative analysis, we compute the layer-wise quantization error as
\begin{equation}
E_q^{(l)}=
\frac{1}{nB}
\sum_{i=1}^{n}
\sum_{b=1}^{B}
\left(
|h_{ib}^{(l)}|-1
\right)^2,
\label{eq:layer}
\end{equation}
where $\mathbf{h}_{i}^{(l)}$ denotes the bounded hash representation obtained from the Hash Register of image $\mathbf{x}_i$ at the $l$-th layer, $h_{ib}^{(l)}$ is its $b$-th dimension, $n$ is the mini-batch size, and $B$ is the hash code length. For the post-quantization baseline, we apply its trained terminal hash projection to the hash-token feature extracted from each layer only for visualization.

As shown in Fig.~\ref{fig:layerwise_quantization}, the Hash Register of HashViT gradually evolves from a continuous distribution to a bimodal distribution concentrated around $-1$ and $+1$ as the depth increases. In contrast, the post-quantization baseline keeps more continuous intermediate distributions and relies on the final projection for code generation. 
Note that the baseline's terminal projection is trained only at the last layer and is applied to intermediate features purely for visualization; under this caveat, the comparison is consistent with binary-friendly representations emerging through the internal evolution of the register in HashViT.

\begin{figure}[t]
\centering
\includegraphics[width=\linewidth]{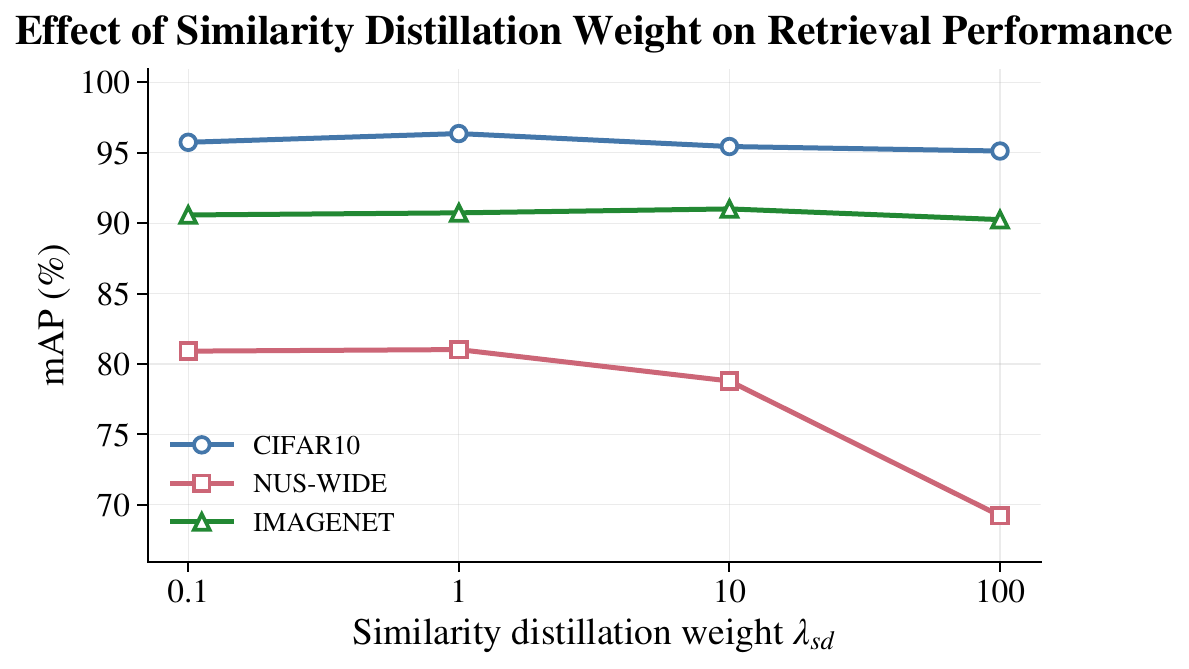}
\caption{Parameter sensitivity analysis of the similarity distillation weight $\lambda_{sd}$ on CIFAR-10, NUS-WIDE, and IMAGENET. The quantization weight $\lambda_q$ is fixed to 0 to isolate the effect of similarity distillation.}
\label{fig:sensitivity_lambda_sd}
\end{figure}

\begin{figure}[t]
\centering
\includegraphics[width=\linewidth]{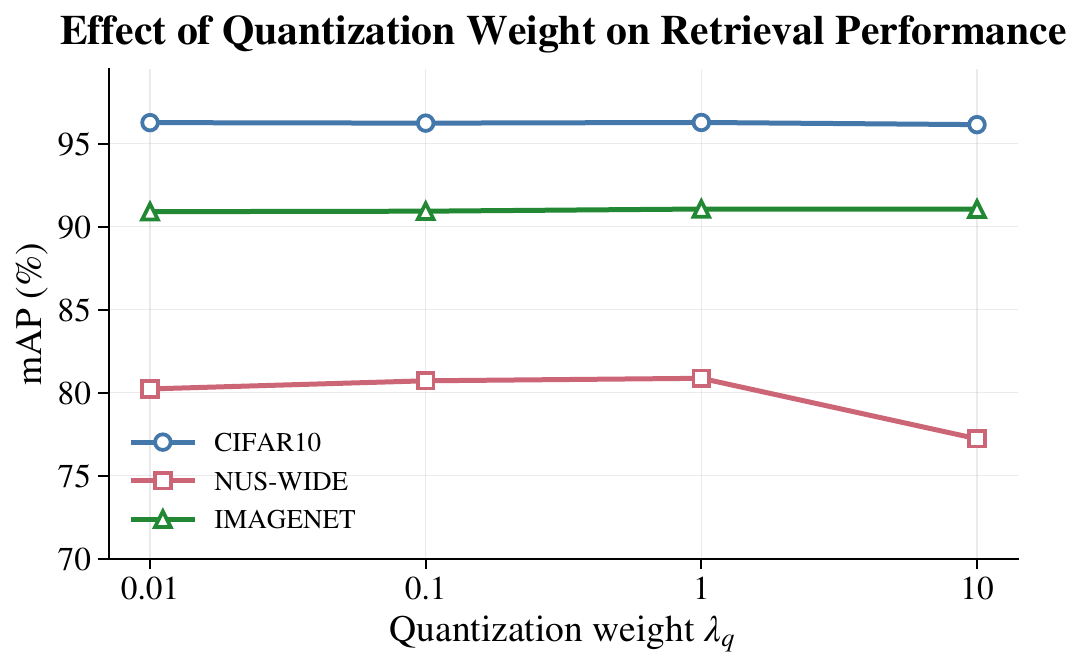}
\caption{Parameter sensitivity analysis of the quantization weight $\lambda_q$ on CIFAR-10, NUS-WIDE, and IMAGENET. For each dataset, $\lambda_{sd}$ is fixed to the best setting selected from the sensitivity analysis.}
\label{fig:sensitivity_lambda_q}
\end{figure}

\subsection{Parameter Sensitivity}
We analyze the sensitivity of HashViT to the loss-weight hyperparameters in the overall objective:
\[
\mathcal{L}
=
\mathcal{L}_{c}
+
\lambda_{sd}\mathcal{L}_{sd}
+
\lambda_{q}\mathcal{L}_{q},
\]
where $\mathcal{L}_{c}$ denotes the learnable semantic hash center loss, $\mathcal{L}_{q}$ is the quantization loss, and $\mathcal{L}_{sd}$ is the similarity distillation loss. Since $\mathcal{L}_{c}$ provides the primary semantic supervision, its weight is fixed to 1. We then study the effects of $\lambda_q$ and $\lambda_{sd}$. Specifically, when evaluating $\lambda_{sd}$, we set $\lambda_q=0$ to isolate the contribution of similarity distillation. When evaluating $\lambda_q$, we fix $\lambda_{sd}$ to the best setting selected for each dataset.

As shown in Fig.~\ref{fig:sensitivity_lambda_sd} and Fig.~\ref{fig:sensitivity_lambda_q}, HashViT exhibits stable performance across a broad range of hyperparameter settings. 

For $\lambda_{sd}$, moderate values generally lead to better results, suggesting that instance-level similarity distillation is beneficial for preserving the semantic topology of CLS representations. However, overly large weights may introduce excessive constraints from the continuous feature space, which is not always aligned with discrete hash code learning. For $\lambda_q$, the performance variation is relatively small, and the best results on CIFAR-10 and NUS-WIDE are obtained even when $\lambda_q=0$. This indicates that HashViT does not require a strong explicit quantization penalty to form binary-friendly representations. Instead, the dedicated HASH token and the native in-transformer learning process already provide a structural tendency toward the binary hash space, while $\mathcal{L}_q$ mainly acts as an auxiliary regularizer on more challenging datasets such as IMAGENET.

Based on the observed trends, we adopt dataset-specific hyperparameter settings in the final experiments. For CIFAR-10 and NUS-WIDE, we set $\lambda_{sd}=1$ and $\lambda_q=0$. For IMAGENET, we set $\lambda_{sd}=10$ and $\lambda_q=1$. These settings consistently achieve the best or near-best performance on the corresponding datasets, demonstrating that HashViT is robust to loss-weight variations and has relatively low hyperparameter sensitivity.

\begin{figure}[t]
\centering
\includegraphics[width=\linewidth]{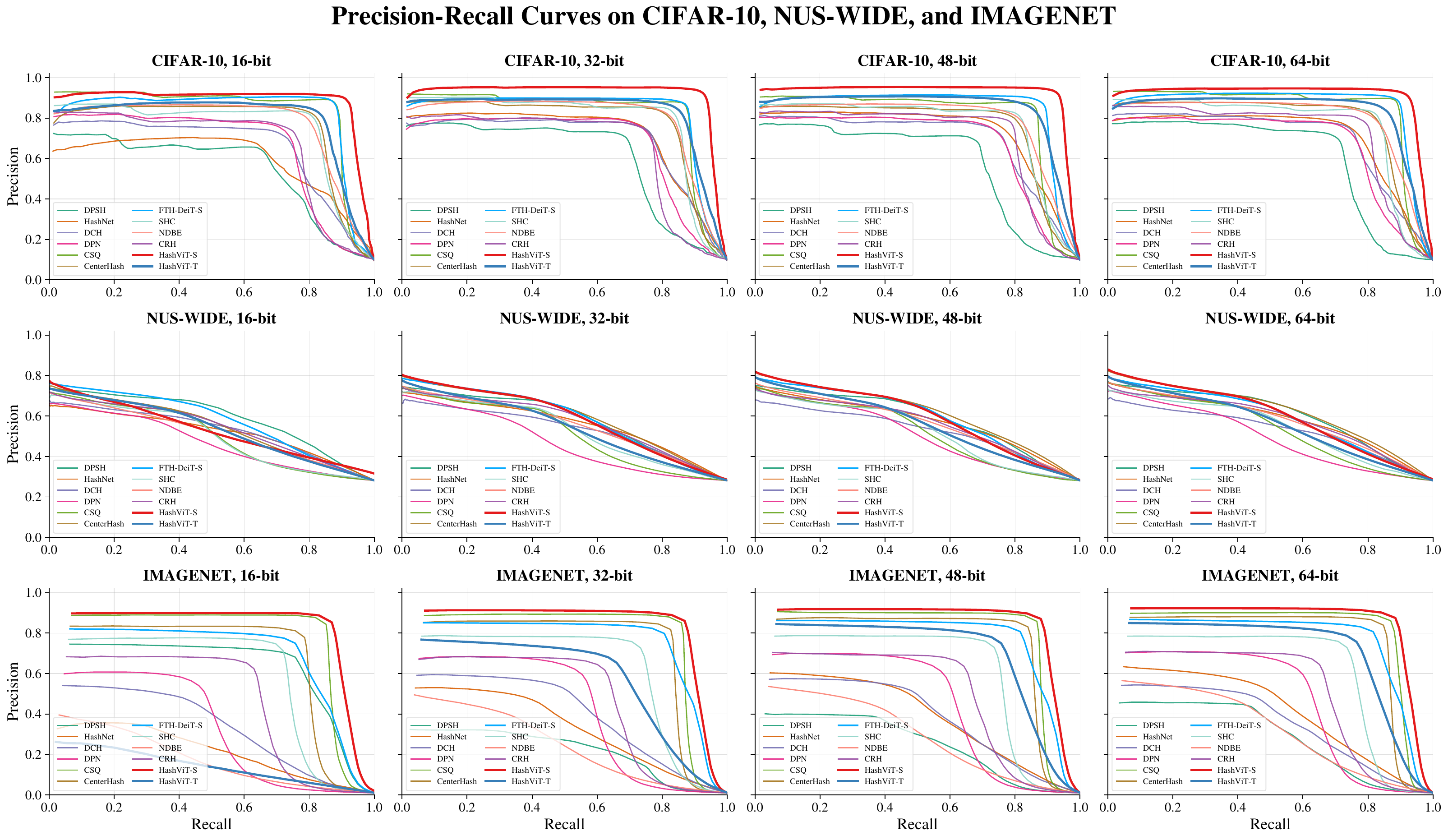}
\caption{Precision-Recall curves on three benchmark datasets. HashViT consistently achieves better precision across different recall levels, indicating stronger semantic preservation in the learned Hamming space.}
\label{fig:pr_curve}
\end{figure}

\begin{figure}[t]
\centering
\includegraphics[width=\linewidth]{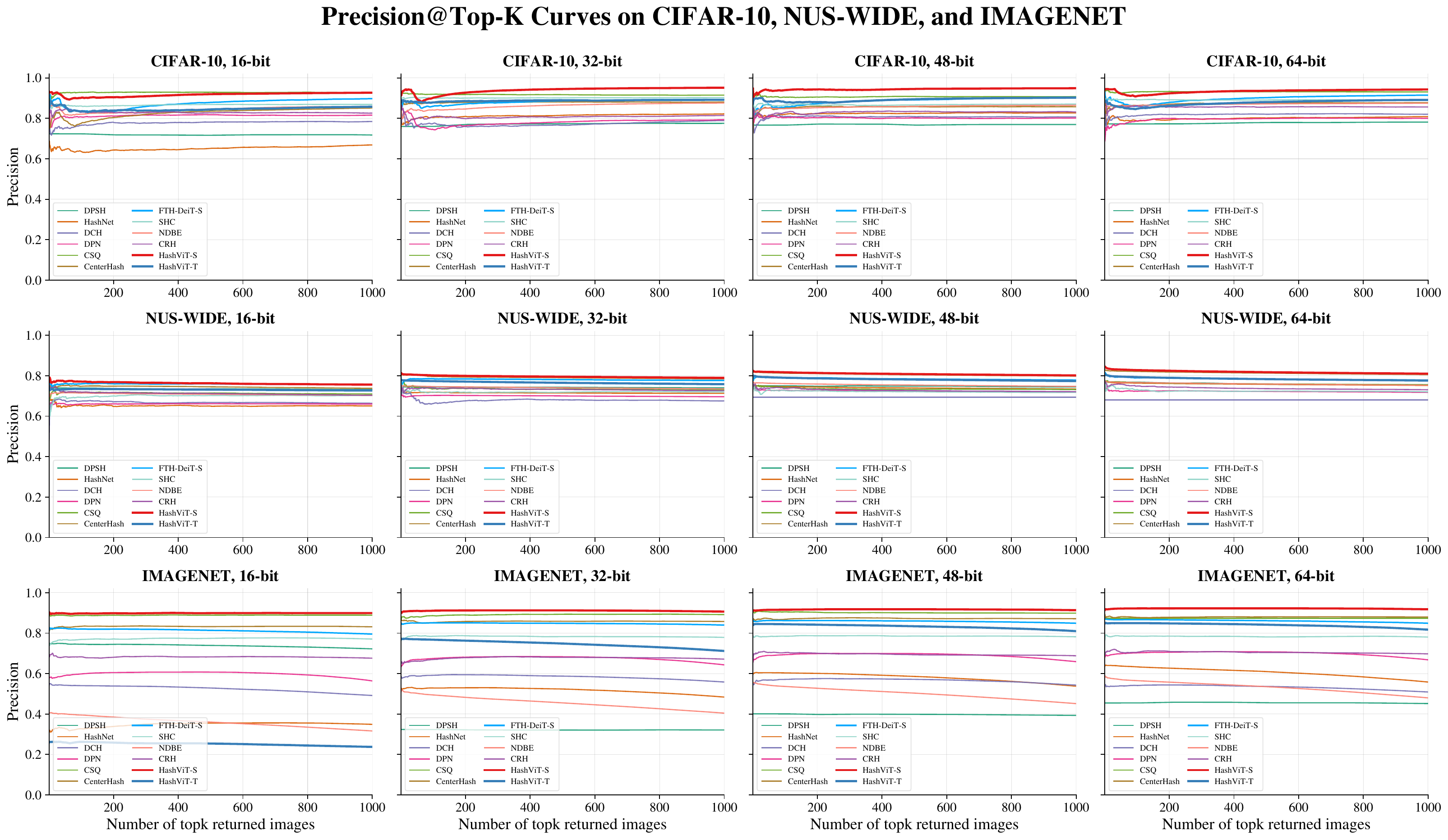}
\caption{Precision@Top-$K$ curves on three benchmark datasets. HashViT obtains higher precision among top-ranked retrieval results, demonstrating its effectiveness in practical image retrieval scenarios.}
\label{fig:p_topk}
\end{figure}

\subsection{Precision-Recall and Precision@Top-$K$ Curves}

We further report the Precision-Recall (PR) curves and Precision@Top-$K$ curves in Fig.~\ref{fig:pr_curve} and Fig.~\ref{fig:p_topk}, respectively. These curves provide complementary evaluations to mAP by showing the retrieval behavior under different recall levels and among top-ranked results. As shown in the figures, HashViT-S consistently achieves superior or highly competitive performance compared with representative baselines. The stronger PR and Precision@Top-$K$ results indicate that the learned binary codes preserve semantic relevance more effectively and improve the quality of top-ranked retrieval results.

\subsection{Efficiency Analysis}
We further analyze the efficiency of HashViT from the perspective of token and parameter overhead. Compared with a standard ViT backbone, HashViT introduces only one additional learnable HASH token. For ViT-Small with $224\times224$ input resolution, the original sequence contains 196 patch tokens and one class token, i.e., 197 tokens in total. After adding the HASH token, the sequence length becomes 198. Therefore, the number of token pairs in self-attention increases only from $197^2$ to $198^2$, corresponding to about $1.0\%$ additional attention computation, while the token-wise projection and MLP overhead is about $0.5\%$. Moreover, HRA is shared across transformer layers and introduces only $B(d-B)+B$ parameters. For the 64-bit setting of ViT-Small, where $d=384$ and $B=64$, this amounts to only $20{,}544$ additional parameters, which is negligible compared with the 21.69M parameters of the backbone.

Despite this minor overhead, HashViT achieves a favorable accuracy-efficiency trade-off. As shown in Table~\ref{tab:hash_comparison}, HashViT-S uses only 21.69M parameters, which is much smaller than ViT-Large-based hashing methods such as TransHash with 292.93M parameters, yet it achieves stronger retrieval performance on CIFAR-10 and ImageNet. Compared with DeiT-based hashing methods, HashViT-S is also more compact than FTH-DeiT-S, which contains 43.06M parameters. These results indicate that the improvement of HashViT does not come from using a heavier backbone, but from introducing a lightweight HASH token and native in-transformer hash learning. Therefore, the additional computational and parameter cost is small relative to the retrieval gain, making HashViT practical for efficient large-scale image retrieval.

\section{Conclusions}
\label{conclusion}
In this paper, we present HashViT, a native hash token learning framework for efficient image retrieval. 
Rather than relying on a terminal hashing layer applied to continuous visual features, HashViT introduces a dedicated HASH token to maintain a hash-oriented retrieval state within the Vision Transformer.
By decomposing the HASH token into a Hash Register and a Semantic Workspace, and refining the register through a lightweight Hash Refinement Adapter, HashViT enables hash representations to evolve progressively across transformer layers while preserving auxiliary continuous semantics.
Extensive experiments on three widely used benchmarks demonstrate that HashViT achieves state-of-the-art or highly competitive retrieval performance with a compact backbone.
Comprehensive ablation studies and layer-wise analyses further confirm the contribution of the dedicated HASH token, the register-workspace decomposition, the Hash Refinement Adapter, and the semantic supervision objective to more discriminative and stable hash representations.
These results suggest that native hash token learning provides an effective alternative to conventional post-quantization hashing for learning compact and discriminative binary representations.




\bibliographystyle{IEEEtran}
\bibliography{sample-base}

@String{Computing = "Computing" }

@String{Computer = "{IEEE} Computer" }

@STRING{aaai    = {Proc.~of the Conf. Artif. Intell. (AAAI)} }

@STRING{aaaiol    = {Proc.~of the Conference on Advancements of Artificial Intelligence (AAAI)} }

@STRING{accv    = {Proc.~of the Asi. Conf. Comput. Vis. (ACCV)} }

@STRING{acmmm=    {Proc.~of the ACM International Conference on Multimedia (ACM MM)} }

@STRING{cvpr    = {Proc. IEEE Conf. Comput. Vis. Pattern Recognit. (CVPR)} }

@STRING{iccv    = {Proc.~of the IEEE/CVF Intl.~Conf.~on Computer Vision (ICCV)} }

@STRING{iclr    = {Proc.~of the Int.~Conf.~on Learning Representations (ICLR)}}

@STRING{icml    = {Proc.~of the Int.~Conf.~on Machine Learning (ICML)} }

@STRING{ijcai   = {Proc.~of the Intl.~Conf.~on Artificial Intelligence (IJCAI)} }

@STRING{neurips = {Proc.~of the Conference on Neural Information Processing Systems (NeurIPS)} }

@STRING{tpami   = {IEEE Trans.~on Pattern Analysis and Machine Intelligence (TPAMI)} }

@STRING{vldb   = {Proc.~of the International Conference on Very Large
               Data Bases (VLDB)} }

@STRING{tmm   = {IEEE Trans.~on Multimedia (TMM)} }

@STRING{icmr   = {Proc.~of international conference on multimedia retrieval (ICMR)} }

@STRING{kbs   = {Knowledge-Based Systems (KBS)} }

@STRING{tcsvt   = {IEEE Transactions on Circuits and Systems for Video Technology (TCSVT)} }

@STRING{tomm = {ACM Transactions on Multimedia Computing, Communications and Applications (TOMM)}}

@STRING{icassp={IEEE International Conference on Acoustics, Speech and Signal Processing (ICASSP)}}

@STRING{spl={IEEE Signal Processing Letters (SPL)}}

@STRING{tois={ACM Transactions on Information Systems (TOIS)}}

@STRING{tnnls={IEEE Transactions on Neural Networks and Learning Systems (TNNLS)}}

@ArtifactSoftware{R,
    title = {R: A Language and Environment for Statistical Computing},
    author = {{R Core Team}},
    organization = {R Foundation for Statistical Computing},
    address = {Vienna, Austria},
    year = {2019},
    url = {https://www.R-project.org/},
}

@inproceedings{adam,
  author    = {Diederik P. Kingma and
               Jimmy Ba},
  title     = {Adam: {A} Method for Stochastic Optimization},
  booktitle = iclr,
  year      = {2015}
}

@inproceedings{pytorch,
  author    = {Adam Paszke and
               Sam Gross and
               Francisco Massa and
               Adam Lerer and
               James Bradbury and
               Gregory Chanan and
               Trevor Killeen and
               Zeming Lin and
               Natalia Gimelshein and
               Luca Antiga and
               Alban Desmaison and
               Andreas K{\"{o}}pf and
               Edward Z. Yang and
               Zachary DeVito and
               Martin Raison and
               Alykhan Tejani and
               Sasank Chilamkurthy and
               Benoit Steiner and
               Lu Fang and
               Junjie Bai and
               Soumith Chintala},
  title     = {PyTorch: An Imperative Style, High-Performance Deep Learning Library},
  booktitle = neurips,
  year      = {2019}
}

@inproceedings{LSH,
  title={Similarity search in high dimensions via hashing},
  author={Gionis, Aristides and Indyk, Piotr and Motwani, Rajeev and others},
  booktitle=vldb,
  year={1999}
}

@article{ITQ,
  author    = {Yunchao Gong and
               Svetlana Lazebnik and
               Albert Gordo and
               Florent Perronnin},
  title     = {Iterative Quantization: {A} Procrustean Approach to Learning Binary
               Codes for Large-Scale Image Retrieval},
  journal   = tpami,
  year      = {2013}
}

@inproceedings{SH,
  author    = {Yair Weiss and
               Antonio Torralba and
               Robert Fergus},
  title     = {Spectral Hashing},
  booktitle = neurips,
  year      = {2008}
}

@inproceedings{DPSH,
  author    = {Wu{-}Jun Li and
               Sheng Wang and
               Wang{-}Cheng Kang},
  title     = {Feature Learning Based Deep Supervised Hashing with Pairwise Labels},
  booktitle = ijcai,
  year      = {2016}
}

@inproceedings{CSQ,
  author    = {Li Yuan and
               Tao Wang and
               Xiaopeng Zhang and
               Francis E. H. Tay and
               Zequn Jie and
               Wei Liu and
               Jiashi Feng},
  title     = {Central Similarity Quantization for Efficient Image and Video Retrieval},
  booktitle = cvpr,
  year      = {2020}
}

@inproceedings{ADSH,
  author    = {Qing{-}Yuan Jiang and
               Wu{-}Jun Li},
  title     = {Asymmetric Deep Supervised Hashing},
  booktitle = aaaiol,
  year      = {2018}
}

@inproceedings{DAPH,
  author    = {Fumin Shen and
               Xin Gao and
               Li Liu and
               Yang Yang and
               Heng Tao Shen},
  title     = {Deep Asymmetric Pairwise Hashing},
  booktitle=  acmmm,
  year      = {2017}
}

@inproceedings{OrthoHash,
  author    = {Jiun Tian Hoe and
               Kam Woh Ng and
               Tianyu Zhang and
               Chee Seng Chan and
               Yi{-}Zhe Song and
               Tao Xiang},
  title     = {One Loss for All: Deep Hashing with a Single Cosine Similarity based
               Learning Objective},
  booktitle = neurips,
  year      = {2021}
}

@inproceedings{HashNet,
  title={Hashnet: Deep learning to hash by continuation},
  author={Cao, Zhangjie and Long, Mingsheng and Wang, Jianmin and Yu, Philip S},
  booktitle=iccv,
  year={2017}
}

@inproceedings{HybridHash,
  title={HybridHash: Hybrid convolutional and self-attention deep hashing for image retrieval},
  author={He, Chao and Wei, Hongxi},
  booktitle=icmr,
  year={2024}
}

@inproceedings{imagenet,
  author    = {Jia Deng and
               Wei Dong and
               Richard Socher and
               Li{-}Jia Li and
               Kai Li and
               Li Fei{-}Fei},
  title     = {ImageNet: {A} large-scale hierarchical image database},
  booktitle = cvpr,
  year      = {2009}
}

@inproceedings{CNNH,
  author    = {Rongkai Xia and
               Yan Pan and
               Hanjiang Lai and
               Cong Liu and
               Shuicheng Yan},
  title     = {Supervised Hashing for Image Retrieval via Image Representation Learning},
  booktitle = aaai,
  year      = {2014},
}

@inproceedings{DCH,
  title={Deep cauchy hashing for hamming space retrieval},
  author={Cao, Yue and Long, Mingsheng and Liu, Bin and Wang, Jianmin},
  booktitle=cvpr,
  year={2018}
}

@inproceedings{pushe,
  title={SHE: Streaming-media Hashing Retrieval},
  author={Pu, Ruitao and Qin, Yang and Song, Xiaomin and Peng, Dezhong and Ren, Zhenwen and Sun, Yuan},
  booktitle=icml,
  year={2025}
}

@article{su2025boundary,
  title={Boundary-aware Prototype Augmentation and Dual-level Knowledge Distillation for Non-Exemplar Class-Incremental Hashing},
  author={Su, Qinghang and Wu, Dayan and Li, Bo},
  journal=kbs,
  year={2025},
  publisher={Elsevier}
}

@article{su2024data,
  title={From data to optimization: Data-free deep incremental hashing with data disambiguation and adaptive proxies},
  author={Su, Qinghang and Wu, Dayan and Wu, Chenming and Li, Bo and Wang, Weiping},
  journal=tcsvt,
  year={2024},
  publisher={IEEE}
}

@inproceedings{wu2024pairwise,
  title={Pairwise-label-based deep incremental hashing with simultaneous code expansion},
  author={Wu, Dayan and Su, Qinghang and Li, Bo and Wang, Weiping},
  booktitle=aaai,
  year={2024}
}

@article{wu2023deep,
  title={Deep uncoupled discrete hashing via similarity matrix decomposition},
  author={Wu, Dayan and Dai, Qi and Li, Bo and Wang, Weiping},
  journal=tomm,
  year={2023},
  publisher={ACM New York, NY}
}

@inproceedings{gu2022deep,
  title={Deep piecewise hashing for efficient hamming space retrieval},
  author={Gu, Jingzi and Wu, Dayan and Fu, Peng and Li, Bo and Wang, Weiping},
  booktitle=icassp,
  year={2022},
}

@inproceedings{wu2019deep,
  title={Deep incremental hashing network for efficient image retrieval},
  author={Wu, Dayan and Dai, Qi and Liu, Jing and Li, Bo and Wang, Weiping},
  booktitle=cvpr,
  year={2019}
}

@inproceedings{yang2020deep,
  title={Deep semantic-alignment hashing for unsupervised cross-modal retrieval},
  author={Yang, Dejie and Wu, Dayan and Zhang, Wanqian and Zhang, Haisu and Li, Bo and Wang, Weiping},
  booktitle=icmr,
  year={2020}
}

@inproceedings{DPN,
  title={Deep Polarized Network for Supervised Learning of Accurate Binary Hashing Codes},
  author={Fan, Lixin and Ng, Kam Woh and Ju, Ce and Zhang, Tianyu and Chan, Chee Seng},
  booktitle=ijcai,
  year={2020}
}

@inproceedings{noh2017large,
  title={Large-scale image retrieval with attentive deep local features},
  author={Noh, Hyeonwoo and Araujo, Andre and Sim, Jack and Weyand, Tobias and Han, Bohyung},
  booktitle=cvpr,
  year={2017}
}

@inproceedings{radenovic2018revisiting,
  title={Revisiting oxford and paris: Large-scale image retrieval benchmarking},
  author={Radenovi{\'c}, Filip and Iscen, Ahmet and Tolias, Giorgos and Avrithis, Yannis and Chum, Ond{\v{r}}ej},
  booktitle=cvpr,
  year={2018}
}

@inproceedings{perronnin2010large,
  title={Large-scale image retrieval with compressed fisher vectors},
  author={Perronnin, Florent and Liu, Yan and S{\'a}nchez, Jorge and Poirier, Herv{\'e}},
  booktitle=cvpr,
  year={2010},
}

@inproceedings{MambaHash,
  title={MambaHash: Visual State Space Deep Hashing Model for Large-Scale Image Retrieval},
  author={He, Chao and Wei, Hongxi},
  booktitle=icmr,
  year={2025}
}

@article{msvit,
  title={Msvit: training multiscale vision transformers for image retrieval},
  author={Li, Xue and Yu, Jiong and Jiang, Shaochen and Lu, Hongchun and Li, Ziyang},
  journal=tmm,
  year={2024},
}

@inproceedings{transhash,
  title={Transhash: Transformer-based hamming hashing for efficient image retrieval},
  author={Chen, Yongbiao and Zhang, Sheng and Liu, Fangxin and Chang, Zhigang and Ye, Mang and Qi, Zhengwei},
  booktitle=icmr,
  year={2022}
}

@inproceedings{FTH,
  title={Factorized Transformer Hashing with Adaptive Routing for Large-scale Image Retrieval},
  author={Huo, Yadong and Qin, Qibing and Zhang, Wenfeng and Huang, Lei and Nie, Jie},
  booktitle=acmmm,
  year={2025}
}

@inproceedings{CenterHash,
  title={Deep hashing with minimal-distance-separated hash centers},
  author={Wang, Liangdao and Pan, Yan and Liu, Cong and Lai, Hanjiang and Yin, Jian and Liu, Ye},
  booktitle=cvpr,
  year={2023}
}

@article{SHC,
  title={Deep hashing with semantic hash centers for image retrieval},
  author={Chen, Li and Liu, Rui and Zhou, Yuxiang and Ma, Xudong and Chen, Yong and Zhang, Dell},
  journal=tois,
  year={2025},
}

@inproceedings{vitregisters,
  title={Vision transformers need registers},
  author={Darcet, Timoth{\'e}e and Oquab, Maxime and Mairal, Julien and Bojanowski, Piotr},
  booktitle=iclr,
  year={2024}
}

@article{LCSQ,
  title={Learnable central similarity quantization for efficient image and video retrieval},
  author={Yuan, Li and Wang, Tao and Zhang, Xiaopeng and Tay, Francis Eng Hock and Jie, Zequn and Tian, Yonghong and Liu, Wei and Feng, Jiashi},
  journal=tnnls,
  year={2024},
}

@inproceedings{HCU,
  title={Deep hashing with hash center update for efficient image retrieval},
  author={Jose, Abin and Filbert, Daniel and Rohlfing, Christian and Ohm, Jens-Rainer},
  booktitle=icassp,
  year={2022},
}

@article{MCR,
  title={Deep hashing with multi-central ranking loss for multi-label image retrieval},
  author={Cui, Can and Huo, Hong and Fang, Tao},
  journal=spl,
  year={2023},
}

@inproceedings{CRH,
  title={Codebook-Centric Deep Hashing: End-to-End Joint Learning of Semantic Hash Centers and Neural Hash Function},
  author={Yin, Shuo and Yin, Zhiyuan and Hou, Yuqing and Liu, Rui and Chen, Yong and Zhang, Dell},
  booktitle=aaai,
  year={2026}
}

@inproceedings{WSHC,
  title={Image retrieval with well-separated semantic hash centers},
  author={Wang, Liangdao and Pan, Yan and Lai, Hanjiang and Yin, Jian},
  booktitle=accv,
  year={2022}
}

@inproceedings{movshovitz2017no,
  title={No fuss distance metric learning using proxies},
  author={Movshovitz-Attias, Yair and Toshev, Alexander and Leung, Thomas K and Ioffe, Sergey and Singh, Saurabh},
  booktitle=iccv,
  year={2017}
}

@inproceedings{kim2020proxy,
  title={Proxy anchor loss for deep metric learning},
  author={Kim, Sungyeon and Kim, Dongwon and Cho, Minsu and Kwak, Suha},
  booktitle={Proceedings of the IEEE/CVF conference on computer vision and pattern recognition},
  pages={3238--3247},
  year={2020}
}

@article{NDBE,
  title={Deep Neighbor Discriminant Binary Embedding for Multi-Label Image Retrieval},
  author={Qin, Qibing and Dou, Mingkun and Zhang, Wenfeng and Huang, Lei and Nie, Jie},
  journal=tmm,
  year={2026},
}

@inproceedings{DCAH,
  title={Discretization Is Not Always Better: Rethinking Deep Quantization for Asymmetric Image Retrieval},
  author={Liu, Xinze and Wu, Dayan and Zhu, Hengjie and Wu, Chenming and Dai, Pengwen},
  booktitle=aaai,
  year={2026}
}

@inproceedings{clip,
  title={Learning transferable visual models from natural language supervision},
  author={Radford, Alec and Kim, Jong Wook and Hallacy, Chris and Ramesh, Aditya and Goh, Gabriel and Agarwal, Sandhini and Sastry, Girish and Askell, Amanda and Mishkin, Pamela and Clark, Jack and others},
  booktitle=icml,
  year={2021},
}





\end{document}